\documentclass{article}

    \PassOptionsToPackage{numbers, compress}{natbib}
 \usepackage[preprint]{neurips_2026}

\usepackage[utf8]{inputenc} 
\usepackage[T1]{fontenc}    
\usepackage{hyperref}       
\usepackage{url}            
\usepackage{booktabs}       
\usepackage{amsfonts}       
\usepackage{nicefrac}       
\usepackage{microtype}      
\usepackage{xcolor}         
\usepackage{graphicx}
\usepackage[table]{xcolor}
\usepackage{multirow}
\usepackage{caption}
\usepackage{wrapfig}
\usepackage{amsmath} 
\usepackage{subcaption}
\usepackage{svg}
\usepackage{algorithmic}
\usepackage{algorithm}

\title{Proprio: Latent Self-Scoring and Inference-Time Refinement for Physically Plausible Video Generation
}

\author{%
  Mariam Hassan $^1$, Kaouther Messaoud $^2$, Wuyang Li $^1$, Alexandre Alahi $^1$  \\
  $^1$ École Polytechnique Fédérale de Lausanne (EPFL), $^2$ Télécom Paris, IP Paris \\
  Project Page: \url{https://vita-epfl.github.io/Proprio/} \\
}

\begin{document}

\maketitle
\vspace{-0.8cm}
\begin{figure}[h]
\label{fig:method}
\centering
\includegraphics[width=\textwidth]{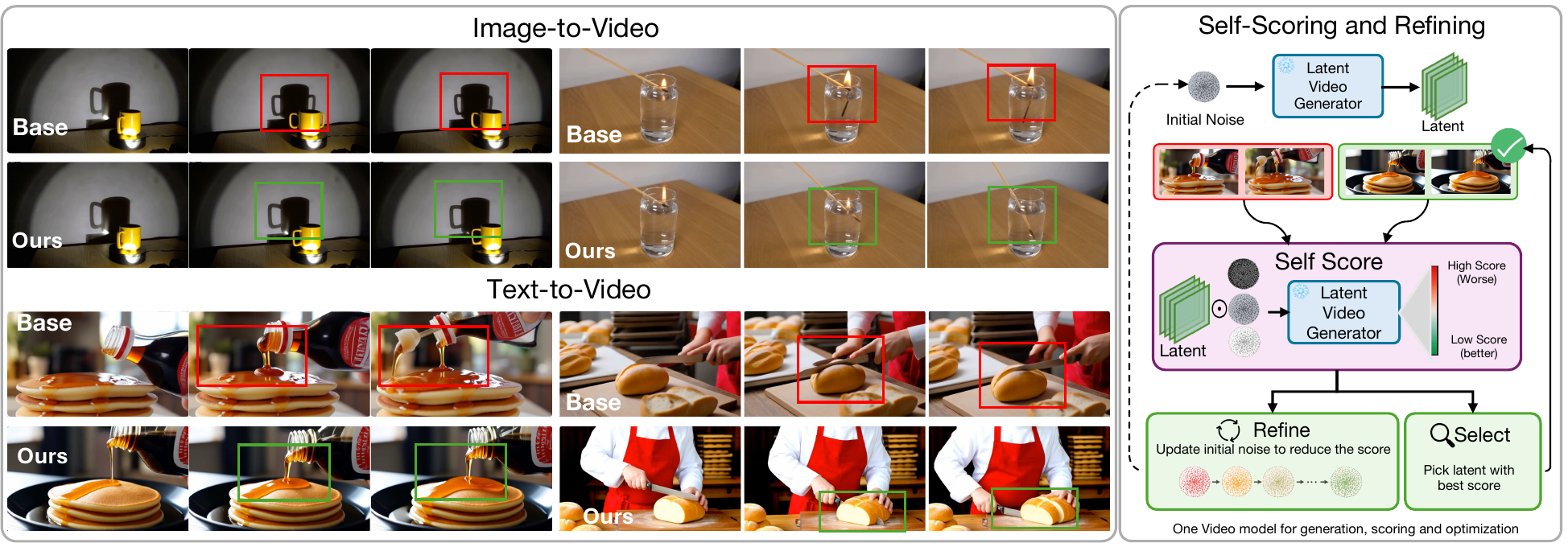}
\caption{\textbf{Proprio enables a video generator to evaluate and improve its own generation.} Left: qualitative refinement comparisons for TurboWan2.2. Right: Proprio computes a latent self-score from the model’s denoising residual and uses it for either sample selection or inference-time refinement. }
\label{fig:pull_figure}
\end{figure}

\definecolor{DiffBlue}{RGB}{242,247,255}
\begin{abstract}
Modern video generative models produce visually impressive results, yet frequently violate basic physical principles. We propose \textbf{Proprio}, a training-free framework that enables a frozen video generator to assess and improve the physical plausibility of its own outputs. Inspired by \textit{proprioception}, the biological sense of one’s own movement, Proprio treats the model's flow residual under controlled latent perturbations as a self-scoring signal. Samples that are better explained by the generator's learned dynamics induce smaller and more stable residuals. We aggregate this signal across timesteps and perturbations, focus it on motion-relevant regions with a dynamic spatiotemporal mask, and use it for best-of-\(N\) search, gradient-based self-refinement, or both. Across text-to-video and image-to-video benchmarks, Proprio consistently improves physical plausibility, outperforming VLM-based scoring, and external world-model baselines in several settings. With TurboWan2.2, Proprio improves Physics-IQ from 32.2 to 37.5 (+16.5\%) and VideoPhy2-hard physical commonsense from 45.6 to 55.0 (+20.6\%). Human evaluation further shows that raters prefer Proprio-selected or refined videos for physical plausibility in roughly two-thirds of comparisons. These results suggest that frozen video generators contain actionable internal signals for evaluating and improving the physical plausibility of their own outputs. 

\end{abstract}

\section{Introduction} \label{sec:intro}

Diffusion models~\cite{ho2020denoising,liu2022flow} have achieved remarkable progress, emerging as dominant paradigms for high-fidelity video generation~\cite{brooks2024sora,ali2025world,wan22}. Despite their impressive visual quality, {\emph{these models still struggle to maintain physical plausibility}}, often producing videos that violate basic real-world constraints. Such failures not only limit perceptual realism but also undermine the utility of these models as reliable world models for downstream applications such as robotics, planning, and simulation.

To address this, existing efforts typically rely on external sources of feedback to provide physical assessment and supervision. Some work inject physical structure through curated datasets, physical priors, or differentiable simulation~\cite{wang2025wisa,liu2024physgen}. Another line of work guides generation using external evaluators, including vision-language models, or pretrained latent world models such as V-JEPA~\cite{yang2025vlipp,cai2025phygdpo,videorepa,wmreward,assran2025v,bardes2024revisiting}. These methods provide useful signals for selecting physically plausible generations and have shown promising results in improving physical realism. 

While effective, this external-feedback paradigm faces two key limitations. \emph{(1) External evaluators introduce their own biases and failures}: VLMs struggle to assess physical plausibility because their judgments are often entangled with text-image alignment objectives and visual appearance~\cite{physicsiq}.  Pretrained world models provide a more dynamics-oriented signal, but operate on decoded videos and map them into their own representation space, creating a representation mismatch with the generator's latent space. \emph{(2) Misalignment with the generator's internal dynamics}: external evaluators are trained with different training objectives, so their feedback is not directly aligned with the generator's sampling process and may miss generator-specific inconsistencies across denoising steps.

These limitations motivate an intrinsic perspective on physical plausibility: \emph{Can a video generator leverage its own denoising dynamics as a self-contained signal for producing more physically plausible videos?} Here, denoising dynamics refers to the model-predicted update direction at each noise level. We focus on its prediction error, which we call the \emph{flow residual}: the discrepancy between the predicted denoising or velocity direction and the target induced by a controlled perturbation.


{\textbf{Conceptually}}, the flow residual can be interpreted as an approximate likelihood signal~\cite{li2023your}, providing inherent cues for assessing physical validity, as previously explored in curated simulated video pairs~\cite{likephys}. In contrast, we reinterpret this likelihood-preference signal as a generator-native objective. Since video generators are trained on large-scale real video datasets, their learned distributions implicitly capture regularities of real-world dynamics. It is therefore natural to treat consistency with such learned dynamics as an intrinsic proxy for physical plausibility. 
{\textbf{Empirically}}, we test this intuition with a small-scale diagnostic check. Given a set of real videos, we pair each video with a generated counterpart that exhibits a physical failure using a video model, and compare their residual scores under that model (see Appendix~\ref{app:diagnostic_natural}). In 82.2\% of the pairs, the residual signal prefers the ground-truth video over the model's own physically implausible generation, suggesting the model's internal denoising dynamics can distinguish natural video dynamics from its own failure cases.

Motivated by this, we propose \emph{\textbf{Proprio}}, a generator-native, training-free framework that converts internal denoising signals into an internal proxy for physical plausibility. Inspired by \emph{\textbf{proprioception}}, the biological ability to sense one’s motion without external observation, Proprio enables a frozen video generator to assess and improve its own outputs. Specifically, we define a \emph{noise-conditioned self-consistency signal} based on the model’s denoising residual under controlled latent perturbations. We robustly estimate the score by aggregating residuals across multiple timesteps with \emph{inverse-variance weighting}, which downweights noisy estimates, and by applying a \emph{dynamic spatiotemporal mask} to emphasize motion-relevant regions where physical failures occur. Beyond scoring, Proprio uses the score as a self-refinement objective. It refines generations by backpropagating through the frozen sampler and updating the initial noise to produce a more internally consistent video.


We evaluate Proprio in three inference-time strategies: best-of-\(N\) search, self-refinement, and hybrid search-and-refine approach (see 
Fig.~\ref{fig:pull_figure}). For image-to-video (I2V) and text-to-video (T2V) generation, Proprio improves physical accuracy over VLM-based, and outperforms external pretrained world-model in several settings. These results suggest that a frozen video generator can be used to generate videos and to guide selection and refinement of its outputs.
Our contributions are as follows:

\begin{itemize}

    \item \textbf{Generator-native self-scoring:} We introduce Proprio, a training-free framework that reuses a frozen generator's denoising residual as an intrinsic score for physical plausibility.

    \item \textbf{Inference-time self-alignment:} We turn this score into an optimization objective for a frozen generator, refining its outputs by updating only the initial noise, and optionally combine this refinement with best-of-\(N\) selection.
   
    \item \textbf{Empirical validation:} We demonstrate consistent gains on I2V and T2V physics benchmarks, showing that a single frozen generator can act as both sampler and self-evaluator, guiding selection and refinement toward more physically plausible dynamics.


\end{itemize}

\section{Related Work} \label{sec:lit}

\textbf{Physical plausibility in video generation.}
Recent video generative models~\cite{brooks2024sora,bar2024lumiere,kondratyuk2024videopoet,wan22,hunyuanvideo,yang2024cogvideox,hacohen2025ltxvideo,moviegen} have achieved impressive visual fidelity, yet they still frequently violate basic physical principles~\cite{kang2025far,physicsiq}. Previous work explored three lines of work to solve problem. The first improves physical fidelity during training by introducing physics-aware data, supervision, or learning objectives~\cite{wang2025wisa,cai2025phygdpo,wang2025wisa}. The second incorporates explicit physical structure or simulation into the generation process~\cite{liu2024physgen}. The final approach addresses the problem at inference time through external guidance, for example using VLM-based planning or reward signals from pretrained world models. VLIPP and GPT4motion use a VLM as physics reward~\cite{yang2025vlipp, lv2024gpt4motion}, while WMReward uses VJEPA-2 as an external latent world model to provide rewards for search and guided sampling~\cite{wmreward}.


\textbf{Inference-time alignment for image generation.}
Inference-time alignment for diffusion models can be categorized into search-based optimization-based methods. Search-based approaches improve alignment by exploring multiple denoising trajectories and selecting or resampling candidates according to a reward, including Best-of-$N$ selection, SMC-style resampling, and value-based decoding~\cite{svdd,dsearch,ma2025inferencescaling,zhang2025inference,jain2025diffusion}. Optimization-based approaches directly modify the sampling trajectory or noise variables to increase the target reward, for example through reward guidance or direct noise optimization~\cite{dno,initno,ye2024tfg,askari2024cvs,dall2025chamfer}. Collectively, these works show that substantial gains can be obtained at test time without retraining the generator. However, they typically rely on externally defined reward models or prompt-alignment objectives.

\textbf{Latent reward modeling.}
Recent work has explored training video generation models as latent reward models, motivated by the limitations of VLM-based evaluators. In particular, learning rewards models by directly training on noisy latent states has been shown to better match latent-space generation and avoid the overhead of pixel-space VLM rewards~\cite{dinalrm,videolrm, nuti2023extracting}. Alternatively, LikePhys~\cite{likephys} curates a synthetic benchmark and evaluates models using preference  Plausibility Preference Error (PPE) metric on curated valid-invalid video pairs. 
Unlike these approaches, which either train dedicated latent reward models or depend on curated simulated pairs, our method directly reuses the frozen generator's own denoising residual as a self-score for physical plausibility.



\begin{figure}[t]
\label{fig:method}
\centering
\includegraphics[width=\textwidth]{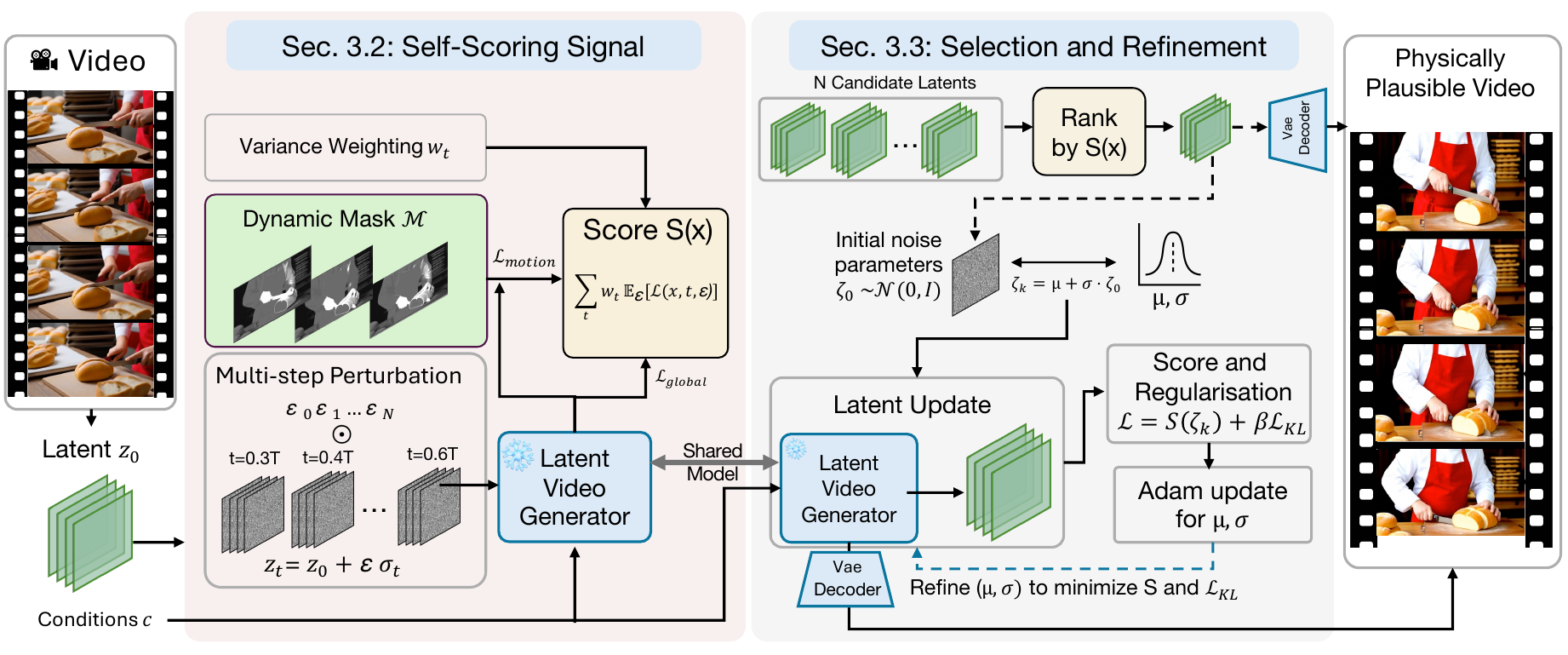}
 \caption{\textbf{Proprio method overview}. For a generated latent video, Proprio computes a per-video self-scoring signal by perturbing the latent at multiple timesteps, measuring denoising residuals with the frozen generator, weighting timesteps by inverse variance, and emphasizing informative motion regions with a dynamic spatiotemporal mask. The resulting score can either be used for ranking and selection, self-refining or both. The VAE encoder is omitted for brevity.}
\label{fig:method}
\end{figure}


\section{Method} \label{sec:method}
We introduce \textbf{Proprio}, a training-free inference-time framework with two complementary stages: self-scoring and self-refinement (see Fig.~\ref{fig:method}). Self-scoring asks whether a generated latent video is consistent with the frozen generator's learned dynamics: samples that the model can explain under controlled perturbations should induce smaller and more stable flow residuals. Since large video generators are trained on natural video dynamics, this learned-flow consistency can serve as an intrinsic proxy for physical plausibility. Self-refinement then uses this score as an objective to improve a sample by updating only a structured parameterization of its initial sampling noise.

The self-scoring signal is built from four components. (1) We perturb each generated latent at multiple scheduler timesteps and noise realizations, probing the sample across different noise regimes. (2) We measure the model's flow prediction residual, reusing the generator's own training objective as a self-consistency signal. (3) We optionally apply a dynamic spatiotemporal mask to focus the score on motion-relevant regions. (4) We aggregate residuals across timesteps with inverse-variance weighting, giving more weight to reliable estimates and downweighting perturbation-sensitive ones. The resulting score supports three inference-time modes: best-of-\(N\) search, self-refinement, and hybrid search-and-refine. We briefly review the flow-matching formulation in Appendix~\ref{app:flow_prelim}.

\subsection{Flow Prediction under Scheduler Perturbations}

Let $c$ denote the conditioning input (e.g., text prompt or input image). Let $z_0 \in \mathbb{R}^{T \times H \times W \times C}$ denote the latent representation of a generated video $x_0$. We consider a pretrained video generator with parameters $\theta$, which predicts a velocity (flow) field
\begin{equation}
\hat{v}_\theta : \mathbb{R}^{T \times H \times W \times C} \times \mathcal{T} \times \mathcal{C} \to \mathbb{R}^{T \times H \times W \times C},
\end{equation}
where $\mathcal{T}$ denotes the set of scheduler timesteps.

Modern video generators define perturbations through a scheduler rather than a fixed analytical path. Given a timestep $t \in \mathcal{T}$ and noise $\epsilon \sim \mathcal{N}(0,I)$, we construct a perturbed latent
\begin{equation}
z_t = \mathcal{P}(z_0, \epsilon, t),
\end{equation}
where $\mathcal{P}$ denotes the forward perturbation process.

The model predicts a flow field $\hat{v}_\theta(z_t, t, c)$, which is trained to match a target transport of the form
\begin{equation}
v^\star(z_0, \epsilon) \approx \epsilon - z_0.
\end{equation}
This corresponds to the standard velocity parameterization and enforces local consistency of the learned transport under perturbations of real data.

\subsection{Intrinsic Self-Scoring via Flow Residuals}

Given a latent sample $z_0$, we probe its consistency by sampling $\epsilon$ and evaluating the model at perturbed states $z_t$. We define the flow residual
\begin{equation}
r(z_0, \epsilon, t; c)
=
\hat{v}_\theta(z_t, t, c) - v^\star(z_0, \epsilon),
\end{equation}
and the per-timestep score
\begin{equation}
\ell_t(z_0, \epsilon; c)
=
\|r(z_0, \epsilon, t; c)\|_2^2.
\end{equation}

This quantity corresponds to the flow-matching regression error evaluated at $z_0$, providing a local surrogate for alignment with the learned data distribution. Intuitively, it measures how consistently the learned transport explains the sample under perturbations: samples near well-modeled regions yield small residuals, while deviations from the learned distribution produce larger discrepancies.

\textbf{Dynamic Spatiotemporal Masking.}
The score $\ell_t$ aggregates errors across the entire latent volume, including large static regions that are less informative for physical plausibility. To focus the signal on informative motion regions, we introduce a dynamic spatiotemporal mask $M \in [0,1]^{T \times H \times W}$.

Let $u$ index spatiotemporal locations. Refer to Appendix~\ref{app:masking} for mask details. The masked score is
\begin{equation}
\ell_t^{\mathrm{motion}}(z_0, \epsilon; c)
=
\frac{\sum_{u} M_u\,\|r_u(z_0,\epsilon,t;c)\|_2^2}
{\sum_{u} M_u + \delta},
\end{equation}
where $\delta > 0$ ensures numerical stability.

This can be interpreted as importance weighting over spatiotemporal regions, emphasizing locations with higher expected relevance to motion and physical interactions, thereby improving sensitivity to physically meaningful deviations.

\textbf{Multi-Perturbation, Multi-Timestep Estimation.}
A single evaluation of $\ell_t$ is sensitive to both timestep and noise. We therefore aggregate across multiple perturbations and timesteps.

Let $\mathcal{T} = \{t_1, \dots, t_K\}$ and $\epsilon^{(n)} \sim \mathcal{N}(0,I)$ for $n=1,\dots,N$. For each timestep $t_k$, we compute
\begin{equation}
\mu_k
=
\frac{1}{N}
\sum_{n=1}^N
\ell_{t_k}(z_0, \epsilon^{(n)}; c),
\end{equation}
\begin{equation}
v_k
=
\frac{1}{N}
\sum_{n=1}^N
\left(\ell_{t_k}(z_0, \epsilon^{(n)}; c) - \mu_k \right)^2.
\end{equation}

The mean \(\mu_k\) estimates the residual strength at timestep \(t_k\), while the variance \(v_k\) measures how stable that estimate is across perturbations. This separates the magnitude of the inconsistency from the reliability of the measurement.

\textbf{Variance-Weighted Aggregation.}
Different timesteps can provide signals of different reliability: some noise levels expose meaningful inconsistencies, while others produce residuals that are highly sensitive to the sampled perturbation. We therefore weight timestep estimates by inverse variance:
\begin{equation}
w_k = \frac{1}{v_k + \varepsilon},
\quad
\tilde{w}_k = \frac{w_k}{\sum_{j=1}^K w_j},
\end{equation}
where \(\varepsilon > 0\) is a small constant for numerical stability.

We use \(v_k\) to measure the stability of the score at timestep \(t_k\): larger \(v_k\) indicates that the score is more sensitive to the perturbation noise. Inverse-variance weighting therefore assigns higher weight to more stable timesteps and lower weight to noisier ones.


The final score is
\begin{equation}
S(z_0; c)
=
\sum_{k=1}^K \tilde{w}_k \,\mu_k.
\end{equation}

For the motion-aware variant, we define \(S_{\mathrm{motion}}\) analogously using \(\ell_t^{\mathrm{motion}}\), and optionally combine both:
\begin{equation}
S_{\mathrm{hybrid}}(z_0;c)
=
\lambda S_{\mathrm{global}}(z_0;c) + (1-\lambda) S_{\mathrm{motion}}(z_0;c),
\end{equation}
where \(\lambda \in [0,1]\).
The global score preserves broad sample consistency, while the motion-aware score emphasizes dynamic regions; the hybrid score balances these two views.





\subsection{Inference-Time Selection and Refinement}
The Proprio score defines a generator-native objective over samples. It can therefore be used not only for selection, but also for refinement: rather than simply choosing the lowest-scoring candidate, we can directly optimize the sampling noise to reduce the score. We thus propose an inference-time self-refinement procedure that updates the initial noise while keeping the generator fixed.

The Proprio score defines a generator-native objective over samples. Given candidates $\{z_0^{(i)}\}$, we select
\(\hat{i}=\arg\min_i S(z_0^{(i)};c)\). This provides a simple test-time scaling mechanism: generate multiple candidates and keep the one that is most internally consistent according to the frozen generator.

Beyond selection, the same score can be used for self-refinement. Instead of modifying the generator weights, we optimize a structured parameterization of the initial sampling noise. Let \(\zeta_0\sim\mathcal{N}(0,I)\) denote the base noise. We introduce per-channel parameters \(\mu\in\mathbb{R}^C\) and \(\sigma\in\mathbb{R}_+^C\), and define
\begin{equation}
\zeta = \mu + \sigma \odot \zeta_0,
\end{equation}
where $\mu \in \mathbb{R}^C$ and $\sigma \in \mathbb{R}_+^C$. We denote by $z_0(\zeta;c)$ the latent generated from noise $\zeta$ under conditioning $c$.

We optimize
\begin{equation}
\mathcal{L}(\mu,\sigma)
=
S(z_0(\zeta;c);c)
+
\beta \mathcal{L}_{\mathrm{KL}},
\end{equation}
where \(\mathcal{L}_{\mathrm{KL}}=\frac{1}{C}\sum_{d=1}^{C}\frac{1}{2}\left(\mu_d^2+\sigma_d^2-1-2\log\sigma_d\right)\).

The Gaussian-affine parameterization gives refinement enough flexibility to improve the sample while keeping the optimized noise close to the generator's original Gaussian prior. Starting from \(\mu=\mathbf{0}\) and \(\sigma=\mathbf{1}\), we optimize \((\mu,\log\sigma)\) with Adam for a small number of steps while keeping the generator fixed. In practice, we return the sample achieving the lowest unregularized Proprio score during optimization. The full algorithm is provided in Appendix~\ref{app:opt}.



\section{Experiments} \label{sec:exp}

We evaluate Proprio in three inference-time settings: \textit{search}, \textit{refinement}, and \textit{search-and-refine} (Sec.~\ref{sec:method}). Experiments are conducted on both text-to-video (T2V) and image-to-video (I2V) generation, using VideoPhy2~\cite{videophy2} and Physics-IQ~\cite{physicsiq}, respectively, two benchmarks designed to assess physical plausibility and reasoning in generated videos. We include additional experiments in Appendix~\ref{app:exp}.

\textbf{Benchmarks} For T2V, we use VideoPhy2~\cite{videophy2}, which reports three metrics: Semantic Adherence (SA), Physical Commonsense (PC), and Joint (SA=1 and PC=1). We follow the default evaluation protocol where scores greater than or equal to 4 are mapped to 1 and all others to 0. Following previous works~\cite{jang2026self}, we randomly sample 360 prompts equally across standard and hard set. For I2V, we use Physics-IQ~\cite{physicsiq}, which directly reports a physics score.

\textbf{Baselines.}
For \textit{search}, we compare against (i) random selection, (ii) a VLM-based scorer using Qwen-7B~\cite{qwen3}, and (iii) WMReward~\cite{wmreward}, which ranks samples using VJEPA2 surprise scores. We follow their official implementation, scoring full video frames and selecting the lowest surprise-score sample using Vjepa2-G. For \textit{refinement}, we compare against WMReward’s guidance-based optimization method using the recommended guidance of 0.005 and update frequency of 5.

\textbf{Implementation details.}
Unless otherwise stated, we generate 16 candidate samples per prompt. For fair comparison, all methods score the same candidate set. Experiments are run on 2 H100 GPUs. We use each model’s default frame rate: 24 FPS for Wan2.2 and Hunyuan Video 1.5, and 16 FPS for Turbo Wan2.2. Refer to Appendix~\ref{app:impl} for more details.

\subsection{Improving Generation via Inference-Time Search}

We evaluate inference-time scaling by generating 16 candidate samples and selecting the best one using Proprio. We conduct experiments across three video models, Wan2.2 5B~\cite{wan22}, TurboDiffusion Wan2.2 14B~\cite{turbo}, and Hunyuan Video 1.5~\cite{hunyuanvideo}, in both (T2V) and (I2V) settings.

\begin{table*}[t]
\centering
\scriptsize
\setlength{\tabcolsep}{2.6pt}
\caption{Inference-time search on VideoPhy2.  We report Semantic Adherence (SA), Physical Commonsense (PC), and Joint accuracy on the standard and hard splits. Higher is better.}
\resizebox{\linewidth}{!}{%
\begin{tabular}{lcccccccccccccccccc}

\toprule

& \multicolumn{6}{c}{\textbf{Wan2.2 5B}} 
& \multicolumn{6}{c}{\textbf{Turbo Wan14B}} 
& \multicolumn{6}{c}{\textbf{Hunyuan Video}} \\

& \multicolumn{3}{c}{Standard} & \multicolumn{3}{c}{Hard}
& \multicolumn{3}{c}{Standard} & \multicolumn{3}{c}{Hard}
& \multicolumn{3}{c}{Standard} & \multicolumn{3}{c}{Hard} \\

\cmidrule(lr){2-4}
\cmidrule(lr){5-7}
\cmidrule(lr){8-10}
\cmidrule(lr){11-13}
\cmidrule(lr){14-16}
\cmidrule(l){17-19}

\textbf{Method}
& SA & PC & Joint & SA & PC & Joint
& SA & PC & Joint & SA & PC & Joint
& SA & PC & Joint & SA & PC & Joint \\

\cmidrule(l){1-19}

Random
& 65.4 & 58.1 & 48.6 & 31.7 & 44.4 & 19.4
& 67.2 & 68.3 & 56.1 & 32.2 & 47.8 & 23.3
& 61.1 & 75.0 & 50.0 & 29.4 & 61.1 & 20.0 \\

Qwen 7B~\cite{qwen3}
& 67.0 & 63.7 & 52.0 & 35.0 & 42.2 & 18.9
& 65.6 & 69.4 & 55.0 & 32.8 & 46.1 & 22.2
& 61.7 & 78.3 & \textbf{52.8} & 31.7 & 64.4 & 23.9  \\

WMReward~\cite{wmreward}
& 62.2 & 66.1 & 51.1 & 33.9 & 47.8 & 20.6
& 63.9 & 73.3 & 53.9 & 30.0 & 53.9 & 20.0
& 57.2 & 78.3 & 48.3 & 27.2 & 70.0 & 21.7 \\

\cmidrule(r){1-19}

\rowcolor{DiffBlue}
Ours - $S_{\mathrm{global}}$
& 67.6 & 64.8 & 52.5 & 30.0 & 50.0 & 18.9
& 65.9 & 73.7 & \textbf{56.4} & 33.3 & 53.3 & 22.2
& 56.7 & 77.2 & 46.1 &  25.0 & 70.8 & 18.8  \\

\rowcolor{DiffBlue}
Ours - $S_{\mathrm{motion}}$
& 66.5 & 68.2 & \textbf{55.3} & 34.4 & 51.7 & \textbf{23.3}
& 64.8 & 74.3 & 55.9 & 33.9 & 57.8 & \textbf{23.9}
& 58.3 & \textbf{79.4} & 50.6 &22.9 & 70.8 & 18.8  \\

\rowcolor{DiffBlue}
Ours - $S_{\mathrm{hybrid}}$
& 66.5 & \textbf{69.8} & \textbf{55.3} & 33.3 & \textbf{52.8} & 22.8
& 64.2 & \textbf{74.9} & \textbf{56.4} & 33.3 & \textbf{58.3} & \textbf{23.9}
& 58.9 & 78.3 & 50.6 & 28.9 & \textbf{71.7} & \textbf{25.0} \\


\bottomrule

\end{tabular}%
}
\label{tab:search_resultst2v}
\end{table*}

\textbf{Text-to-Video.}
As shown in Table~\ref{tab:search_resultst2v}, Proprio consistently outperforms random selection and the VLM-based scorer on VideoPhy2, while WMReward remains a stronger baseline than Qwen-7B. Across models, Qwen-7B provides limited benefit: although it occasionally improves semantic adherence (SA), these gains do not translate into stronger physical commonsense (PC) or joint performance, consistent with prior observations~\cite{wmreward}. Compared to WMReward, Proprio achieves further improvements in most settings, particularly on PC and joint metrics. We observe a clear tradeoff between SA and PC: global scoring tends to better preserve SA, whereas motion-aware and hybrid scoring more strongly improve physical consistency. Among these, the hybrid variant is the most robust, consistently achieving the best balance between SA and PC and the strongest joint performance. While improving physical plausibility can reduce SA, Proprio exhibits a milder tradeoff than prior approaches, with smaller SA degradation relative to the gains in PC and joint scores. Overall, these results indicate that generator-native self-scoring provides a more effective signal for selecting physically plausible T2V samples than generic VLM-based evaluation, and is competitive with or stronger than WMReward.

\begin{wraptable}{r}{0.48\linewidth}
\vspace{-10pt}
\centering
\scriptsize
\setlength{\tabcolsep}{3pt}
\caption{Inference-time search results on Physics-IQ across three models. Higher is better.}
\label{tab:search_resultsi2v}
\begin{tabular}{lccc}
\toprule
\textbf{Method} & \textbf{Wan2.2 5B} & \textbf{Turbo} & \textbf{Hunyuan Video} \\
\midrule
Random & 29.45 & 31.25 & 31.63 \\
Qwen 7B~\cite{qwen3} & 28.51 & 31.32 & 32.23 \\
WMReward~\cite{wmreward} & 32.40 & 34.67 & 34.75 \\
\midrule
\rowcolor{DiffBlue}
Ours - $S_{\mathrm{global}}$ & 31.47 & 34.48 & 33.73 \\
\rowcolor{DiffBlue}
Ours - $S_{\mathrm{motion}}$ & 32.25 & 34.65 & \textbf{36.09} \\
\rowcolor{DiffBlue}
Ours - $S_{\mathrm{hybrid}}$& \textbf{32.46} & \textbf{35.13} & 35.13 \\
\bottomrule
\end{tabular}
\vspace{-10pt}
\end{wraptable}

\textbf{Image-to-Video.}
As shown in Table~\ref{tab:search_resultsi2v}, the I2V setting exhibits a different trend. The motion and hybrid variants of Proprio perform similarly, with motion-based scoring slightly stronger sometimes. This is consistent with the fact that I2V candidates are anchored to the same input image and are therefore already semantically aligned, making motion dynamics the primary factors distinguishing samples. While WMReward performs strongly across models, Proprio slightly improves over it on Wan2.2 5B and Turbo and achieves the best overall result on Hunyuan Video with the motion-aware score.  Overall, these results show that generator-native self-scoring provides a strong and competitive criterion for selecting physically plausible I2V generations.

\subsection{Refining Generations with Inference-Time Optimization}
We evaluate inference-time refinement using Proprio as the optimization signal. To make gradient-based optimization tractable, we use a distilled model (TurboDiffusion Wan 14B) and backpropagate through a reduced number of denoising steps (4 instead of 50), significantly lowering computational cost while preserving a useful refinement signal. We conduct experiments in both text-to-video (T2V) and image-to-video (I2V) settings on the VideoPhy2~\cite{videophy2} and Physics-IQ~\cite{physicsiq} benchmarks.

As shown in Table~\ref{tab:opt_unified}, Proprio consistently improves over the base model and outperforms WMReward, with the largest gains observed when combining search and refinement. Refinement alone already yields meaningful improvements, and initializing from better candidates (via search) and then refining them leads to consistently stronger results, indicating that Proprio benefits from both selection and optimization. On Physics-IQ, the motion variant is the strongest, meaning that motion-focused refinement is very effective for I2V physical plausibility. On VideoPhy2, the hybrid score typically yields the strongest physical consistency on the standard split, while both motion and hybrid variants substantially improve performance on the hard split. Across settings, the most consistent gains are observed on physical commonsense (PC), indicating that refinement primarily improves physical realism. Overall, these results demonstrate that Proprio provides an effective optimization signal for challenging physical reasoning tasks, with motion-aware scoring particularly beneficial for I2V and hybrid scoring offering the most robust improvements for T2V.

\begin{table*}[t]
\centering
\small
\setlength{\tabcolsep}{4pt}
\caption{Inference-time refinement on Turbo Wan2.2. We report Physics-IQ and the VideoPhy2 breakdown on the standard and hard subsets. Higher is better.}
\label{tab:opt_unified}
\begin{tabular}{l c ccc ccc}
\toprule

& \multicolumn{1}{c}{\textbf{Physics-IQ}} & \multicolumn{6}{c}{\textbf{VideoPhy2}} \\
\textbf{Method} &  & \multicolumn{3}{c}{\textbf{Standard}} & \multicolumn{3}{c}{\textbf{Hard}} \\
& Score & SA & PC & Joint & SA & PC & Joint \\
\midrule

Turbo Wan2.2
& 32.2
& 65.6 & 67.2 & 55.0
& 31.1 & 45.6 & 19.4 \\

\cmidrule(r){1-8}

\multicolumn{8}{c}{\textit{Refinement only}} \\
\quad + WMReward ($\mathcal{G}$)$^\ast$~\cite{wmreward}
& 32.4
& 67.3 & 67.2 & 55.6
& 32.8 & 45.0 & 21.7 \\
\rowcolor{DiffBlue}
\quad + Proprio ($\nabla$ $S_{\mathrm{global}}$)
& 34.6
& 65.4 & 68.2 & 54.2
& 29.4 & 43.9 & 20.0 \\
\rowcolor{DiffBlue}
\quad + Proprio ($\nabla$ $S_{\mathrm{motion}}$)
& \textbf{35.8}
& 65.4 & 69.3 & 54.2
& 32.2 & 47.2 & 20.6 \\
\rowcolor{DiffBlue}
\quad + Proprio ($\nabla$ $S_{\mathrm{hybrid}}$)
& 34.9
& 65.6 & \textbf{71.7} & 54.4
& 32.2 & \textbf{50.6} & 21.1 \\

\cmidrule(r){1-8}

\multicolumn{8}{c}{\textit{Search  (BoN4) + refinement}} \\
\quad + WMReward ($\mathcal{G}$ + BoN4)$^\ast$~\cite{wmreward}
& 34.7
& 64.4 & 67.2 & 51.1
& 32.8 & 48.9 & 19.4 \\
\rowcolor{DiffBlue}
\quad + Proprio ($\nabla$ $S_{\mathrm{global}}$ + BoN4)
& 36.7
& 64.4 & 71.7 & 54.4
& 29.4 & 54.4 & 21.1 \\
\rowcolor{DiffBlue}
\quad + Proprio ($\nabla$ $S_{\mathrm{motion}}$ + BoN4)
& \textbf{37.4}
& 66.1 & 71.7 & 52.8
& 29.7 & 54.3 & 21.1 \\
\rowcolor{DiffBlue}
\quad + Proprio ($\nabla$ $S_{\mathrm{hybrid}}$ + BoN4)
& 37.1
& 63.9 & \textbf{73.3} & 51.1
& 31.1 & \textbf{55.0} & 21.1 \\

\cmidrule(r){1-8}

\multicolumn{8}{c}{\textit{Search (BoN8) + refinement }} \\
\quad + WMReward ($\mathcal{G}$ + BoN8)$^\ast$~\cite{wmreward}
& 35.5
& 66.7 & 68.9 & 52.8
& 30.6 & 52.2 & 21.1 \\
\rowcolor{DiffBlue}
\quad + Proprio ($\nabla$ $S_{\mathrm{motion}}$  + BoN8)
& \textbf{37.5}
& 65.0 & 73.3 & 55.6
& 31.7 & \textbf{54.4} & 23.3 \\
\rowcolor{DiffBlue}
\quad + Proprio ($\nabla$ $S_{\mathrm{hybrid}}$  + BoN8)
& 37.3
& 66.1 & \textbf{74.4} & 57.8
& 29.6 & 54.2 & 20.7 \\

\bottomrule
\end{tabular}
\\
\tiny $^\ast$ self reproduced. $\mathcal{G}$ indicates guidance
\end{table*}

\subsection{Human evaluation} \label{sec:human_study}
\begin{wrapfigure}{r}{0.41\textwidth}
    \vspace{-12pt}
    \centering
    \includegraphics[width=0.4\textwidth, trim={0 0.1cm 0 1.0cm}, clip]{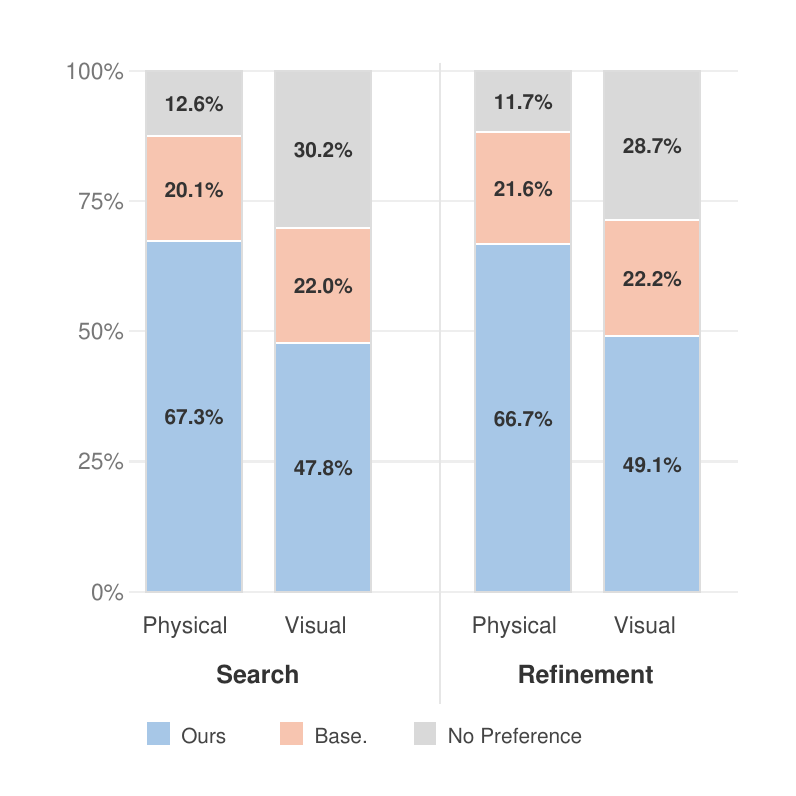}
    \caption{Human preference assessing visual quality and physical plausibility. }
    \label{fig:human_eval}
    \vspace{-11pt}
\end{wrapfigure}
To complement benchmark-based evaluation, we conduct a pairwise human evaluation study to test whether Proprio's self-score agrees with human judgments. We sample 64 video pairs for search and refinement settings for both T2V and I2V, and collect 330 responses. In the search setting, each pair compares a low-score and high-score Proprio sample for the same condition; in refinement, each pair compares a Proprio-refined video with its unrefined baseline. As shown in Fig.~\ref{fig:human_eval}, human preferences align well with Proprio on physical plausibility. Raters prefer the lower-score sample in 67.3\% of search comparisons and the Proprio-refined sample in 66.7\% of refinement comparisons. The effect is weaker for visual quality, where Proprio-selected or refined videos are preferred in roughly half of comparisons and the remaining votes are split between the baseline and no preference. This suggests that Proprio primarily improves perceived physical plausibility while largely preserving visual quality. Additional preference results and visual quality evaluation using VBench~\cite{huang2023vbench} are provided in Appendix~\ref{app:exp}.

\section{Ablation Studies}~\label{sec:ablation}
\textbf{Effect of noise levels for scoring.} \label{sec:ablation_noise}
As shown in Fig.~\ref{fig:noise_ablation} and Table~\ref{tab:ablation_var_midnoise}, the most informative scoring signal arises from an intermediate noise regime, rather than from very low or very high noise levels. On Physics-IQ, both global $S_{\mathrm{global}}$ and motion aware $S_{\mathrm{motion}}$  scoring improve substantially when evaluation is restricted to mid-noise timesteps, while performance degrades at the extremes. A similar pattern holds on VideoPhy2: aggregating scores over the full noise range yields the highest semantic adherence (SA), but significantly weaker physical commonsense (PC) and joint performance. In contrast, restricting evaluation to mid-noise timesteps improves PC and produces the strongest joint scores. These results indicate that extreme noise levels yield less reliable or less informative signals, whereas mid-noise perturbations provide the most discriminative signal for physical plausibility.

\textbf{Effect of variance weighting.}
Table~\ref{tab:ablation_var_midnoise} shows that inverse-variance weighting improves score aggregation by emphasizing timesteps with more stable residual estimates. This effect is most pronounced for the motion score, where variance weighting increases PC from 66.48 to 68.16, yielding the best overall PC result. Gains for the global score are smaller but consistent, indicating that not all timesteps contribute equally reliable information. These results highlight that effective self-scoring depends not only on selecting an appropriate noise regime, but also on weighting timesteps according to their stability. In practice, combining mid-noise with variance-weighted motion scoring provides the strongest and most robust performance. We note that this effect is significantly weaker on distilled models, where variance weighting yields only marginal improvements (see Appendix~\ref{app:var_turbo}).

\begin{wraptable}{r}{0.50\linewidth}
\vspace{-10pt}
\centering
\small
\setlength{\tabcolsep}{4pt}
\caption{Dynamic masking strategies for optimization on Turbo Wan2.2.}
\label{tab:masking_ablation}
\begin{tabular}{l c ccc}
\toprule
& \textbf{Physics-IQ} & \multicolumn{3}{c}{\textbf{VideoPhy2 Standard}} \\
\textbf{Masking} & \textbf{Score} & \textbf{SA} & \textbf{PC} & \textbf{Joint} \\
\midrule
Motion & \textbf{35.8} & 65.36 & \textbf{69.27} & \textbf{54.19} \\
Attention-based & 34.10 & 63.33 & 68.33 & 52.22 \\
Textaware & 33.04 & \textbf{66.66} & 65.56 & 53.33 \\
\bottomrule
\end{tabular}
\vspace{-8pt}
\end{wraptable}
\textbf{Dynamic masking strategy.} We explore different setups for the motion-based masking.
Table~\ref{tab:masking_ablation} compares three dynamic masking variants for optimization. The \emph{motion} mask is the default introduced in Sec.~\ref{sec:method}. The \emph{textaware} mask derives motion from cross-attention, making it explicitly prompt-grounded, while the \emph{attention-based} mask uses self-attention feature changes as an internal proxy for motion, similar to~\cite{pondaven2025video}. Additional details are provided in Appendix~\ref{app:masking}. All three variants provide useful dynamic signals, but exhibit distinct tradeoffs. The motion mask achieves the strongest overall performance, with the best Physics-IQ score and the highest PC and joint scores on VideoPhy2, indicating that direct motion cues are the most reliable signal in this setting. The textaware mask yields the highest semantic adherence (SA), suggesting that prompt-grounded masking better preserves semantic alignment, but at some cost to physical consistency. The attention-based variant remains competitive but is slightly weaker. Overall, these results show that Proprio is compatible with multiple forms of dynamic masking, and that the choice of mask controls the tradeoff between semantic adherence and physical consistency.

\begin{figure*}[t]
\centering

\begin{minipage}[b]{0.45\textwidth}
\centering
\includegraphics[width=\linewidth,trim={0 5 0 5},clip]{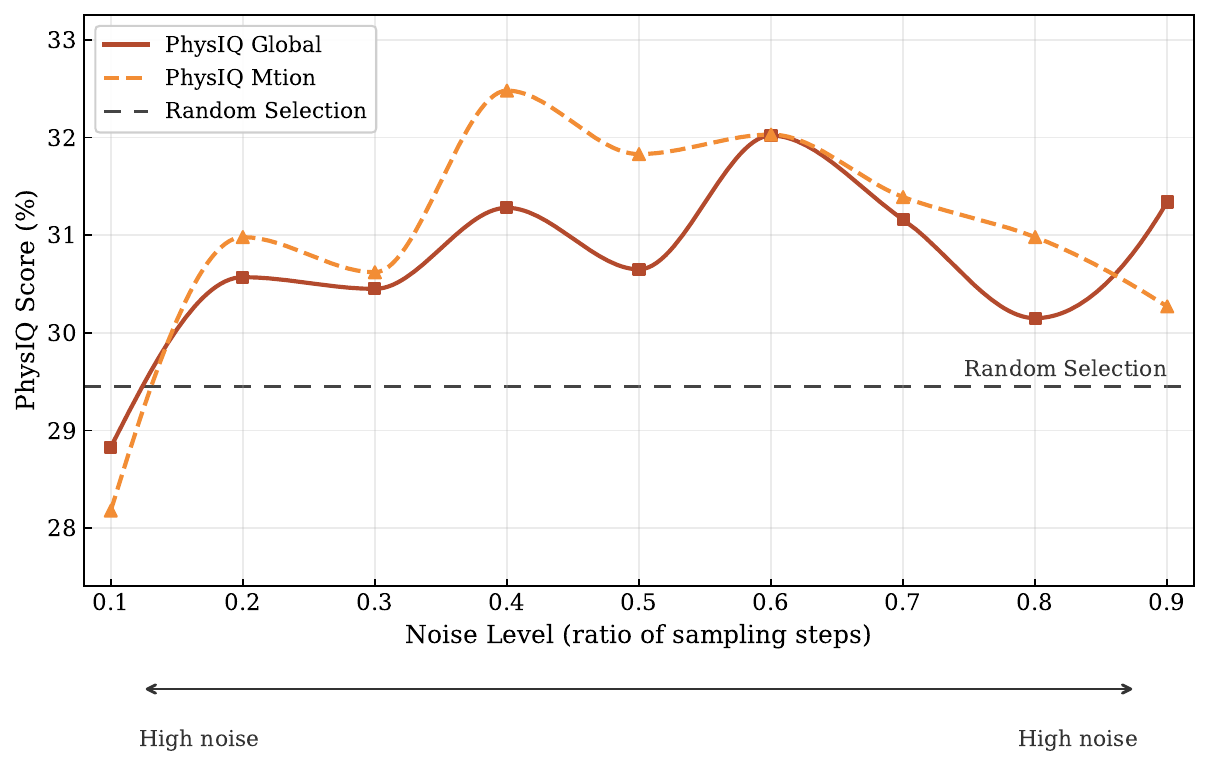}
\caption{Effect of noise levels at different timestep ranges on Physics-IQ for Wan2.2 5b.}

\label{fig:noise_ablation}
\end{minipage}
\hfill
\begin{minipage}[b]{0.52\textwidth}
\centering
\small
\setlength{\tabcolsep}{4pt}
\begin{tabular}{llccc}
\toprule
 & & \multicolumn{3}{c}{\textbf{VideoPhy2}} \\
\textbf{Range} & \textbf{Method} & \textbf{SA} & \textbf{PC} & \textbf{Joint} \\
\midrule
\multirow{2}{*}{\begin{tabular}[c]{@{}c@{}}All noise \\ ($0.2$--$0.8$)\end{tabular}} 
& $S_{\mathrm{global}}$ & 71.66 & 60.56 & 51.66 \\
& $S_{\mathrm{motion}}$ & 69.44 & 61.66 & 51.11 \\
\midrule
\multirow{4}{*}{\begin{tabular}[c]{@{}c@{}}Mid noise \\ ($0.3$--$0.6$)\end{tabular}}
& $S_{\mathrm{global}}$ & 67.04 & 64.25 & 51.40 \\
& $S_{\mathrm{motion}}$ & 67.03 & 66.48 & 56.42 \\
& $S_{\mathrm{global}}$ + var & 67.59 & 64.80 & 52.51 \\
& $S_{\mathrm{motion}}$ + var & 66.48 & \textbf{68.16} & 55.31 \\

\bottomrule
\end{tabular}
\captionof{table}{Effect of aggregating scores at different noise levels and with variance weighting on scoring for VideoPhy2 (standard) with Wan2.2 5B.}
\label{tab:ablation_var_midnoise}
\end{minipage}

\end{figure*}
\section{Discussion}

\textbf{Importance of mid-level noise.}
Our ablations in Section~\ref{sec:ablation_noise} show that the most informative self-scoring signal arises at intermediate noise levels. At very low noise, the perturbed latent remains too close to the original sample, making the residual largely local and less sensitive to higher-level violations of physical plausibility. At very high noise, much of the sample-specific structure is lost, so predictions are driven more by the model’s generic prior and become less discriminative. Mid-level noise provides the best balance: it preserves enough structure for the score to remain sample-specific while perturbing the latent enough to reveal inconsistencies in spatiotemporal dynamics. This explains the stronger PC and joint performance of mid-noise evaluation relative to low or high-noise regimes.

\textbf{Inference scaling versus inference refinement.}
Comparing search and refinement reveals a clear dependence on the generation setting. In T2V, search is particularly effective because samples from the same prompt exhibit substantial scene and motion diversity, making selection from a larger candidate pool highly beneficial. This aligns with prior works~\cite{fvg} discussing the diversity-performance tradeoff in T2V. In contrast, I2V candidates are anchored to the same input image, reducing semantic diversity and leaving motion quality as the primary source of variation. As a result, the advantage of large-scale search diminishes, and the gap between search and refinement narrows. In this regime, motion-aware refinement becomes effective even with a smaller search budget. Overall, these results indicate that scaling search is more important for T2V, whereas in I2V the marginal benefit of additional diversity is lower and refinement plays a stronger role.


\section{Conclusion and Limitations} \label{sec:conc}
We introduced \textbf{Proprio}, a generator-native inference-time framework for improving physical plausibility in video generation. By reusing the frozen generator’s own denoising dynamics as an intrinsic self-scoring signal, Proprio enables both search and refinement without external reward models. Across text-to-video and image-to-video benchmarks, Proprio consistently improves physical plausibility and provides a strong alternative to external evaluators. Proprio is inherently limited by the knowledge encoded in the base generator: it cannot recover dynamics that the model has not learned. In addition, like other inference-time scaling methods, Proprio increases inference cost by requiring additional candidate scoring, search, or refinement steps. Our refinement experiments also rely on distilled models for computational tractability, and scaling gradient-based refinement to larger, non-distilled generators remains challenging due to the cost of backpropagation through long denoising trajectories. An important direction for future work is therefore the development of more efficient scoring and refinement. Overall, our results suggest a shift in perspective for video generation: rather than relying solely on external supervision or auxiliary reward models, future systems may increasingly exploit generator-internal signals for self-evaluation and self-alignment.

\section{Acknowledgments}
This work was supported as part of the Swiss AI Initiative by a grant from the Swiss National Supercomputing Centre (CSCS) under project ID a144. Kaouther Messaoud is supported by Hi! PARIS and ANR/France 2030 program.

{
    \small
    \bibliographystyle{ieeenat_fullname}
    \bibliography{main}
}

\newpage
\appendix

\section{Additional experiments and ablations} \label{app:exp}

\subsection{Diagnostic Study: Preference for natural video dynamics} \label{app:diagnostic_natural}
To test whether the self-score captures a useful preference for natural video dynamics, we conduct a small but highly curated diagnostic check on Physics-IQ. We select 45 \textit{distinct} scenarios from Physics-IQ and use their ground-truth videos as physically plausible references. For each case, we construct a matched physical-failure example using Turbo Wan2.2, either by prompting the model toward a specific implausible behavior or by generating multiple samples and selecting one with a clear physics violation. We manually curate these pairs so that the generated counterpart remains visually and semantically close to the reference where possible, while the main intended difference is physical plausibility. The physical failures include, unrealistic collisons, rigid objects changing shapes, unrealistic shadows, liquid dynamics, unrealistic falls, wrong reaction to magnet and more.
We then score both the ground-truth video and the generated failure using the same Turbo Wan2.2 model with the motion Proprio loss, and compare which video is preferred by the self-score. 

\begin{figure}[H]
\centering
\includegraphics[width=\textwidth]{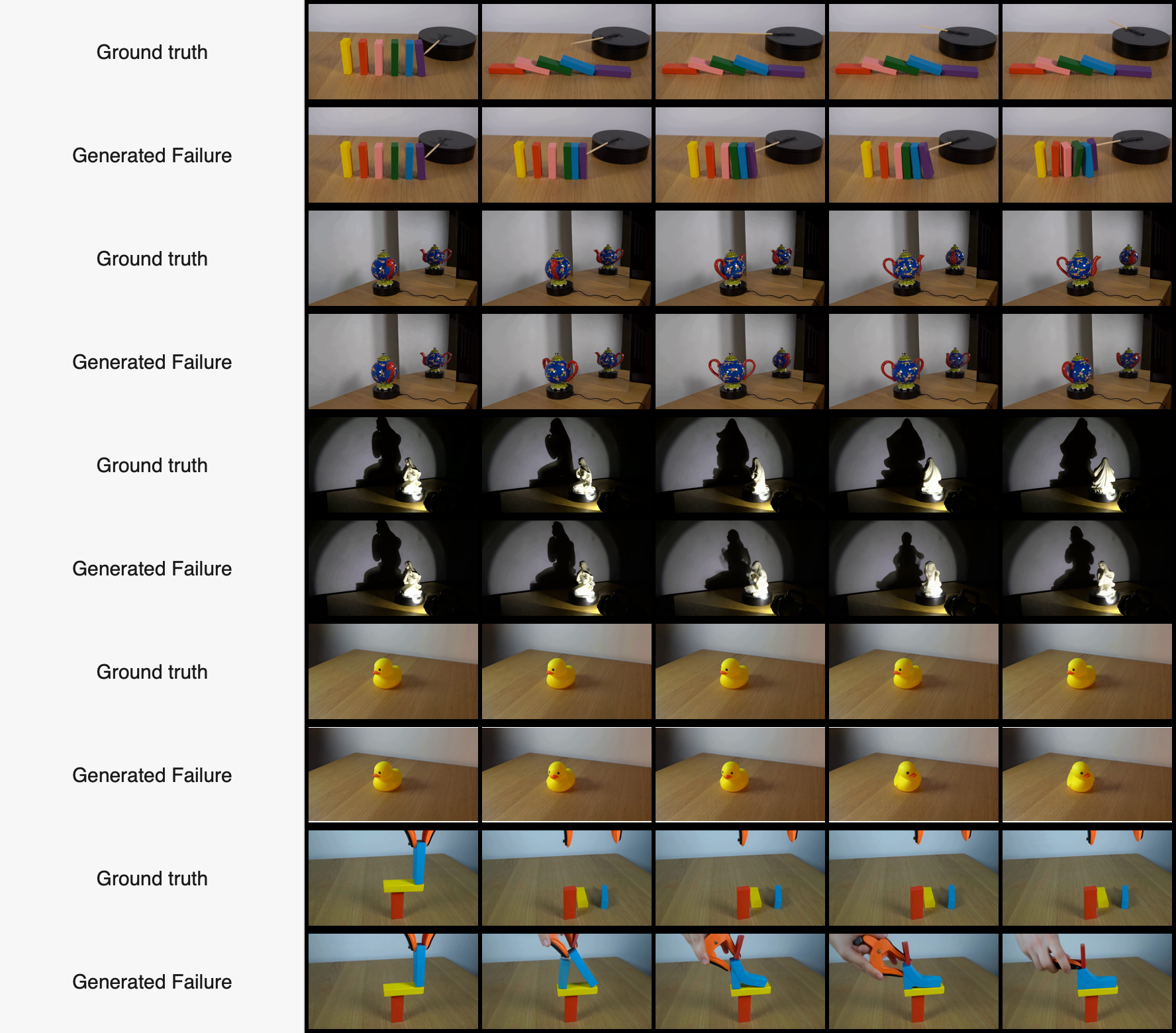}
\caption{\textbf{Correctly preferred diagnostic examples.} Each pair shows a ground-truth Physics-IQ video and a generated Turbo Wan2.2 sample with a visible physical failure. These examples are drawn from the subset where the dynamic Proprio self-score correctly prefers the ground-truth video, illustrating that the generator-native residual can favor natural video dynamics over physically implausible generations.}
\label{fig:study1}
\end{figure}

\begin{figure}[H]
\centering
\includegraphics[width=\textwidth]{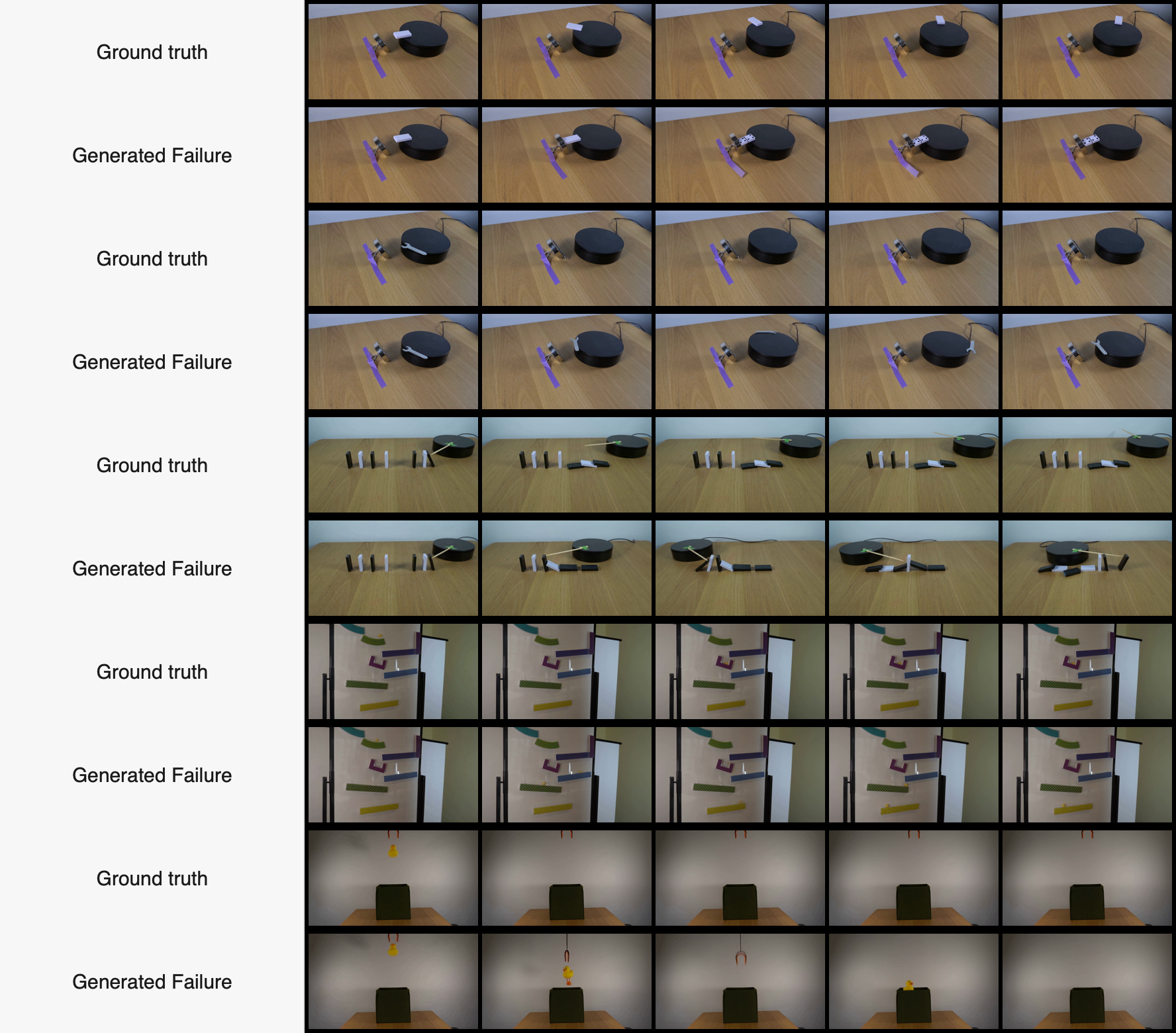}
 \caption{\textbf{Correctly preferred diagnostic examples.} Each pair shows a ground-truth Physics-IQ video and a generated Turbo Wan2.2 sample with a visible physical failure. These examples are drawn from the subset where the Proprio self-score correctly prefers the ground-truth video, illustrating that the generator-native residual can favor natural video dynamics over its own physically implausible generations.}
\label{fig:study2}
\end{figure}

The ground-truth video is preferred in 82.2\% of pairs (37/45), suggesting that the generator-native residual tends to favor natural video dynamics over the model’s own physically implausible generations. This study is intended only as a diagnostic sanity check supporting our interpretation that learned-distribution consistency can serve as a useful proxy for physical plausibility. We show examples of the curated pairs in Figures~\ref{fig:study1} and~\ref{fig:study2}.

\subsection{Visual Quality} \label{app:exp_quality}

To assess whether the gains in physical plausibility come at the expense of visual quality, we additionally evaluate the generated videos on VBench~\cite{huang2023vbench}. As shown in Table~\ref{tab:vbench_quality}, Proprio does not degrade perceptual quality across either T2V or I2V. Subject consistency, background consistency, motion smoothness, aesthetic quality, and imaging quality remain close to the baseline across refinement, search, and search+refinement settings. In some cases, these metrics slightly improve, suggesting that Proprio's physical-plausibility gains are not obtained by sacrificing visual quality. Overall, these results indicate that the improvements in physical plausibility obtained by Proprio do not come at the cost of perceptual video quality, and may in some cases slightly improve it.

\begin{table*}[h]
\centering
\small
\setlength{\tabcolsep}{5pt}
\caption{VBench quality metrics for Turbo Wan2.2 under baseline sampling, Proprio refinement ($\nabla$), search (BoN), and search + refinement ($\nabla$ + BoN).}
\label{tab:vbench_quality}
\begin{tabular}{lccccc}
\toprule
\textbf{Method} & \textbf{Subject} & \textbf{Background} & \textbf{Motion} & \textbf{Aesthetic} & \textbf{Imaging} \\
& \textbf{Consistency} & \textbf{Consistency} & \textbf{Smoothness} & \textbf{Quality} & \textbf{Quality} \\
\midrule
\multicolumn{6}{c}{\textit{VideoPhy2 (T2V)}} \\
Turbo Wan2.2 & 0.92 & 0.92 & 0.99 & 0.51 & 0.67 \\
\quad + Proprio($\nabla$) & 0.93 & 0.93 & 0.98 & 0.51 & 0.68 \\
\quad + Proprio(BoN) & 0.95 & 0.95 & 0.99 & 0.54 & 0.68 \\
\quad + Proprio($\nabla$ + BoN) & 0.93 & 0.93 & 0.98 & 0.52 & 0.68 \\
\midrule
\multicolumn{6}{c}{\textit{PhysicsIQ (I2V)}} \\
Turbo Wan2.2 & 0.95 & 0.96 & 0.99 & 0.49 & 0.66 \\
\quad + Proprio($\nabla$) & 0.95 & 0.96 & 0.99 & 0.49 & 0.66 \\
\quad + Proprio(BoN) & 0.95 & 0.96 & 0.99 & 0.48 & 0.68 \\
\quad + Proprio($\nabla$ + BoN) & 0.96 & 0.96 & 0.99 & 0.49 & 0.66 \\
\bottomrule
\end{tabular}
\end{table*}

\subsection{Time Comparison} \label{app:time}
Table~\ref{tab:runtime} compares the runtime overhead of Proprio and WMReward relative to baseline generation. Proprio is particularly efficient in I2V, where refinement adds only limited overhead and remains substantially cheaper than WMReward guidance, both with and without search. In T2V, the runtime gap is smaller: Proprio refinement is comparable in cost to WMReward guidance, while Proprio search+refinement becomes the most expensive setting because it combines candidate selection with gradient-based refinement. However, it is important to note that inference-time scaling in general requires an increase in compute to search for the best generation and the time required scales linearly with number of samples used for the search.
\begin{table}[h]
\centering
\small
\setlength{\tabcolsep}{5pt}
\caption{Runtime comparison of Proprio and WMReward in I2V and T2V settings using Turbo Wan. We report average runtime in seconds and relative runtime normalized to the baseline. We use one sample generation for this experiment across 3 runs. Numbers increase linearly with samples.}
\label{tab:runtime}
\begin{tabular}{llcc}
\toprule
\textbf{Domain} & \textbf{Method} & \textbf{Time (s)} & \textbf{$\times$Time} \\
\midrule
\multirow{5}{*}{I2V}
& Baseline & $37.54 \pm 0.06$ & 1.00 \\
& Ours $\nabla$ & $49.11 \pm 0.25$ & 1.31 \\
& Ours (BON + $\nabla$) & $67.59 \pm 0.11$ & 1.80   \\
& WMReward \cite{wmreward} $\mathcal{G}$  & $87.35 \pm 4.46$ & 2.33  \\
& WMReward \cite{wmreward} (BON + $\mathcal{G}$) & $92.24 \pm 0.44$ & 2.46  \\
\midrule
\multirow{5}{*}{T2V}
& Baseline & $14.36 \pm 0.18$ & 1.00 \\
& Ours$\nabla$ & $29.72 \pm 0.05$ & 2.07 \\
& Ours (BON + $\nabla$) & $49.67 \pm 0.06$ & 3.46  \\
& WMReward \cite{wmreward} $\mathcal{G}$ & $30.38 \pm 7.48$ & 2.12 \\
& WMReward \cite{wmreward} (BON + $\mathcal{G}$) & $37.69 \pm 4.63$ & 2.62 \\
\bottomrule
\end{tabular}
\end{table}

\subsection{Effect of variance on distilled model} \label{app:var_turbo}

Table~\ref{tab:var_weight_turbo} shows that variance weighting has only a limited effect on the distilled Turbo model. On Physics-IQ, using variance weighting does not improve performance and in fact leads to slightly lower scores for the global and combined variants, while leaving the dynamic variant unchanged. A similar pattern appears on VideoPhy2 where the differences are small across all three variants, and only minor shifts between semantic adherence and physical consistency. In particular, variance weighting slightly improves PC for the global and combined variants, but these gains are offset by lower SA or joint performance, and the dynamic variant is nearly unchanged. This contrasts with the non-distilled Wan model, where variance weighting provides clearer gains. One possible explanation is that distillation compresses the denoising trajectory into fewer steps, making the timestep-wise residual estimates more uniform in reliability. We therefore, do not use variance weighting for optimization setting as it's done on the distilled model.
\begin{table*}[h]
\centering
\small
\setlength{\tabcolsep}{4pt}
\caption{Effect of variance weighting on the distilled Turbo model.}
\label{tab:var_weight_turbo}
\begin{tabular}{lcc|cccccc}
\toprule
& \multicolumn{2}{c|}{\textbf{Physics-IQ}} & \multicolumn{6}{c}{\textbf{VideoPhy2 Standard}} \\
\cmidrule(lr){2-3}
\cmidrule(lr){4-9}
\textbf{Variant} & \textbf{w/o variance} & \textbf{w/ variance}
& \textbf{SA} & \textbf{PC} & \textbf{Joint}
& \textbf{SA} & \textbf{PC} & \textbf{Joint} \\
&  &  & \multicolumn{3}{c}{\textbf{w/o variance}} & \multicolumn{3}{c}{\textbf{w/ variance}} \\
\midrule
$S$   & 34.5 & 34.3 & 66.7 & 73.3 & 58.9 & 65.9 & 73.7 & 56.4 \\
$S_{\mathrm{motion}}$  & 34.6 & 34.6 & 65.0 & 74.4 & 56.7 & 64.8 & 74.3 & 55.9 \\
$S_{\mathrm{hybrid}}$ & 35.1 & 34.6 & 64.4 & 74.4 & 56.1 & 64.2 & 74.9 & 56.4 \\
\bottomrule
\end{tabular}
\end{table*}

\subsection{Motion-stratified results}
\label{app:motion_control}
While motion is an important aspect of physical plausibility, its magnitude alone is not a proxy for physical correctness. A physically plausible video may contain little motion when the scene demands it, whereas an implausible generation may contain excessive or inconsistent motion. We therefore evaluate whether Proprio improves physical plausibility among samples with comparable motion magnitude. We compute a motion magnitude score per video and split videos into quartiles within each dataset for TurboWan2.2. The low-motion bin contains videos up to the 33rd percentile, the medium-motion bin contains videos between the 33rd and 67th percentiles, and the high-motion bin contains videos above the 67th percentile. This data-adaptive binning avoids arbitrary thresholds and enables fair within-dataset comparisons across motion regimes.
As shown in Table~\ref{tab:motion_stratified_control}, Proprio-selected samples generally improve benchmark performance within motion bins. On Physics-IQ, Proprio outperforms random selection, with particularly large gains in the low- and high-motion regimes. On VideoPhy2, Proprio improves the PC metric in the low- and medium-motion bins and remains comparable in the high-motion bin. These results indicate that Proprio's gains are not explained solely by changes in motion magnitude; instead, the score continues to identify more physically plausible generations among videos with similar motion levels.
\begin{table}[h]
\centering
\small
\setlength{\tabcolsep}{4pt}
\caption{Motion-stratified benchmark control. Within each motion bin, we compare Proprio samples against random samples. For Physics-IQ, we report benchmark score; for VideoPhy2, we report PC metric. Positive $\Delta$ indicates Proprio-selected samples outperform random samples.}
\label{tab:motion_stratified_control}
\begin{tabular}{llccc}
\toprule
\textbf{Benchmark} & \textbf{Motion bin} & \textbf{Proprio} & \textbf{Random} & $\boldsymbol{\Delta}$ \\
\midrule
\multirow{3}{*}{Physics-IQ}
& Low    & \textbf{32.0} & 24.8 & +7.2 \\
& Medium & \textbf{33.0} & \textbf{33.0} & +0.0 \\
& High   & \textbf{41.0} & 33.6 & +7.4 \\
\midrule
\multirow{3}{*}{VideoPhy2}
& Low    & \textbf{81.9} & 70.8 & +11.1 \\
& Medium & \textbf{78.9} & 77.4 & +1.5 \\
& High   & 58.0 & \textbf{59.4} & -1.4 \\
\bottomrule
\end{tabular}
\end{table}

\subsection{Human Evaluation} \label{app:human_study}

\begin{wrapfigure}{r}{0.5\linewidth}

    \vspace{-10pt}

    \centering

    \includegraphics[width=1\linewidth,]{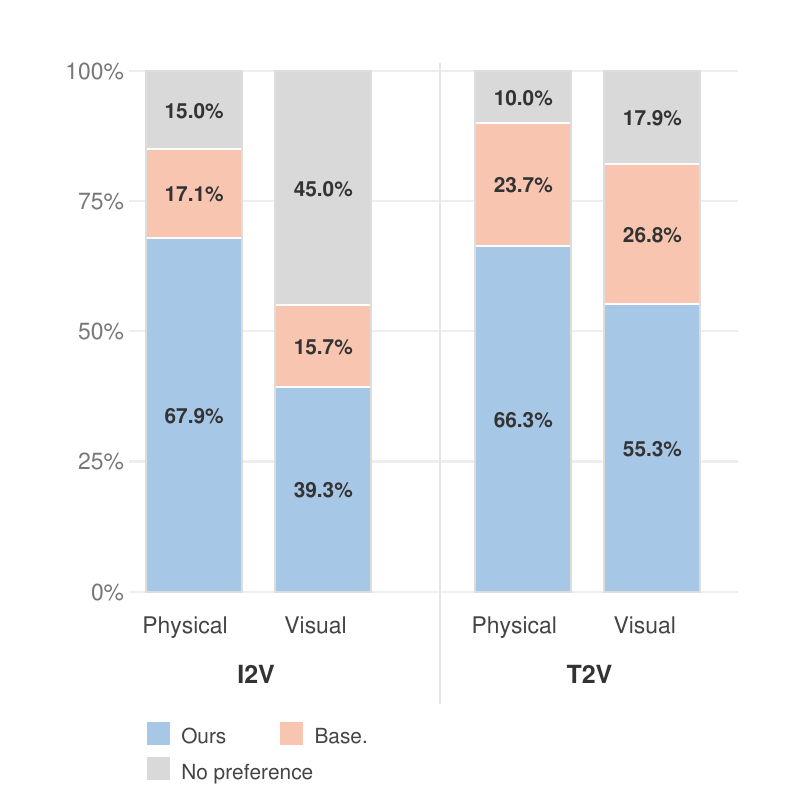}

    \caption{Human preference results split by generation setting.}

    \label{fig:mode-human-study}

    \vspace{-10pt}

\end{wrapfigure}

We conduct a human evaluation with overall results presented in Sec~\ref{sec:human_study} but we provide more results in this section.
Figure~\ref{fig:mode-human-study} provides the human evaluation results split by generation setting. The trend is consistent across both I2V and T2V: voters prefer Proprio for physical plausibility in the majority of comparisons, with 67.9\% preference in I2V and 66.3\% in T2V. This suggests that Proprio's improvements are perceived by humans in both conditional image-to-video and text-to-video settings. For visual quality, the pattern is more mode-dependent. In I2V, most annotators report no preference, which is expected because both videos are anchored by the same input image and therefore often share very similar scene content and appearance. In T2V, visual preferences are more decisive, with 55.3\% favoring Proprio, likely because generated scenes can differ more substantially in terms of composition. Overall, the human study indicates that Proprio's main benefit is improved perceived physical plausibility, while visual quality is generally preserved and sometimes preferred.

\begin{figure}[h]
\centering
\includegraphics[width=\textwidth]{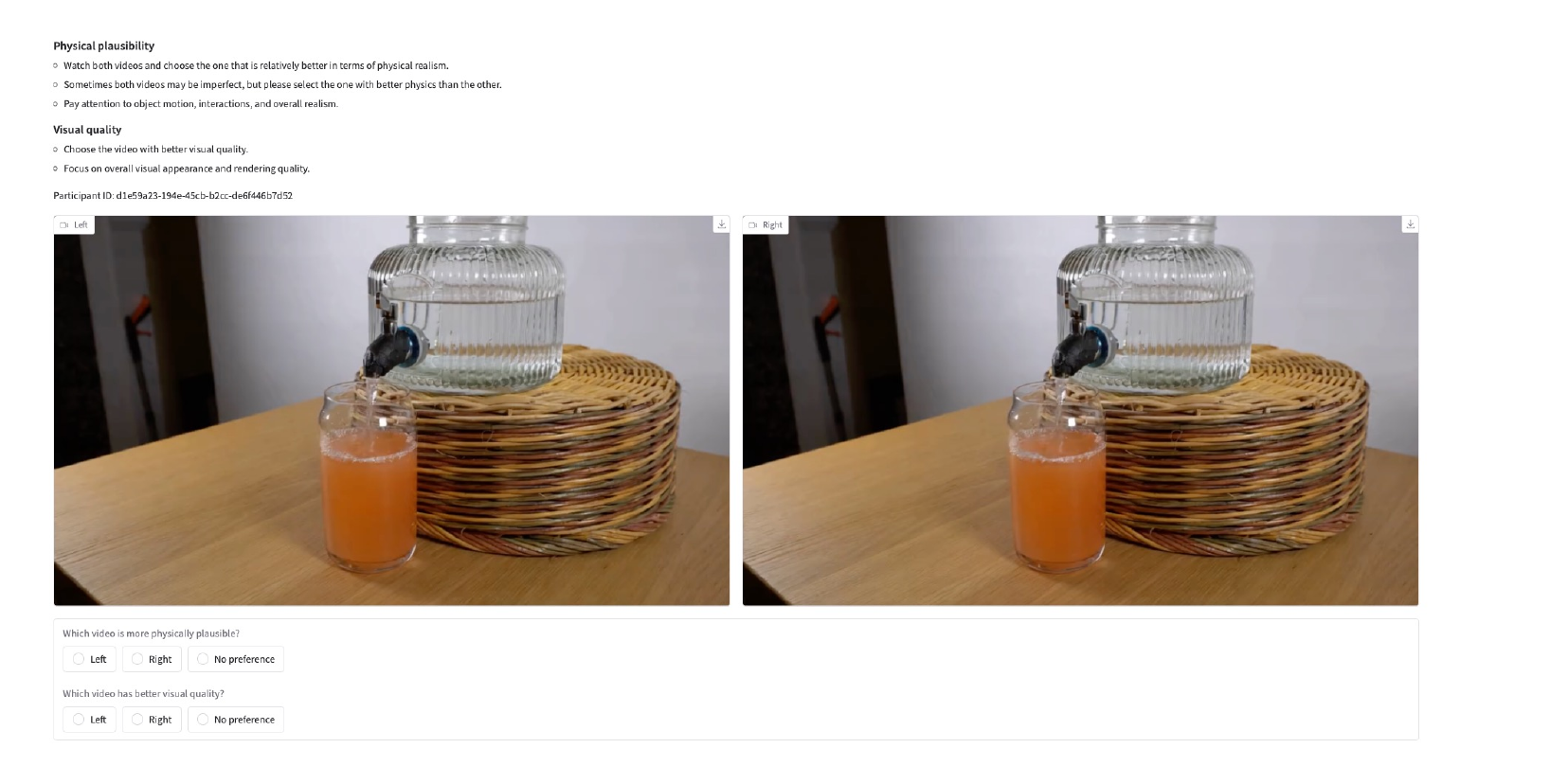}
 \caption{A screenshot from the human evaluation study showing full instructions and layout used.}
\label{fig:study}
\end{figure}

We additionally provide a screenshot example of the study with full instructions given to participants in figure~\ref{fig:study}. Participants volunteered to answer the questions with no compensation. No personal data were collected, only answers to the questions of study. 

\section{Additional Discussion} \label{app:discussion}
\textbf{Semantic Adherence - Physical Commonsense tradeoff.}
Improving physical plausibility in T2V is inherently multi-objective: candidates that faithfully realize complex prompts may involve difficult interactions and therefore be more prone to physical errors, while physically safer generations may simplify the requested action or reduce interaction complexity. This has also been observed in prior works~\cite{wmreward}. While Proprio exhibits similar tradeoff, some preserve semantic adherence better; for example $S_{\mathrm{global}}$ generally preserves SA better while its improvement on physics commonsense is slightly less than $S_{\mathrm{motion}}$ as seen in Section~\ref{sec:exp}. Similarly, the text-aware dynamic mask in Section~\ref{sec:ablation} better preserves semantic adherence, at the cost of slightly weaker physical improvements. In practice, the choice of score can therefore be matched to the application: $S_{\mathrm{global}}$ is preferable when prompt fidelity is critical, $S_{\mathrm{motion}}$ when physical dynamics are the main concern, and $S_{\mathrm{hybrid}}$ as a balanced default for improving physical plausibility while maintaining semantic alignment.

\section{More Details on optimization} \label{app:opt}

Algorithm~\ref{alg:self_refinement} summarizes the self-refinement procedure used in Sec.~\ref{sec:method}. Starting from the original sampling seed, we optimize a Gaussian-affine reparameterization of the initial noise, \(\zeta = \mu + \sigma \odot \zeta_0\), while keeping the video generator fixed. At each iteration, the current noise is used to generate a latent sample \(z_0(\zeta;c)\), which is evaluated with the Proprio score together with a KL regularizer that keeps the refined noise close to the standard Gaussian prior. The optimization is performed over \((\mu,\log \sigma)\) using Adam, and the final output is the sample achieving the lowest unregularized Proprio score during refinement. This parameterization constrains the update to a structured family of perturbations, which stabilizes optimization and reduces drift away from the generator's original sampling distribution.

\begin{algorithm}[h]
\caption{Proprio Self-Refinement Algorithm}
\label{alg:self_refinement}
\begin{algorithmic}[1]
\STATE Sample base noise $\zeta_0 \sim \mathcal{N}(0,I)$ from the initial seed.
\STATE Initialize $\mu \leftarrow \mathbf{0}$ and $\log \sigma \leftarrow \mathbf{0}$.
\FOR{$n=1$ \TO $N$}
    \STATE $\sigma \leftarrow \exp(\log \sigma)$
    \STATE $\zeta \leftarrow \mu + \sigma \odot \zeta_0$
    \STATE Generate latent sample $z_0(\zeta;c)$ with the frozen generator
    \STATE Evaluate Proprio score $S\!\left(z_0(\zeta;c);c\right)$
    \STATE Compute
    \[
    \mathcal{L}_{\mathrm{KL}}
    =
    \frac{1}{C}\sum_{d=1}^{C}
    \frac{1}{2}\left(\mu_d^2+\sigma_d^2-1-2\log\sigma_d\right)
    \]
    \STATE Form the optimization objective
    \[
    \mathcal{L}(\mu,\sigma)
    =
    S\!\left(z_0(\zeta;c);c\right)
    +
    \beta \mathcal{L}_{\mathrm{KL}}
    \]
    \STATE Update $(\mu,\log \sigma)$ with Adam (optional gradient clipping)
\ENDFOR
\STATE \textbf{return} the sample $z_0(\zeta^\star;c)$ achieving the lowest unregularized Proprio score $S\!\left(z_0(\zeta;c);c\right)$ during optimization
\end{algorithmic}
\end{algorithm}

\section{Dynamic masking strategies} \label{app:masking}

We consider three dynamic masking strategies in ablations~\ref{sec:ablation}, each designed to emphasize regions that are most informative for physical plausibility. All three produce a spatiotemporal saliency volume
\(
S \in \mathbb{R}^{T \times H \times W}
\)
that is later converted into the latent dynamic mask used for scoring and refinement. The variants differ only in how this saliency volume is estimated.

\paragraph{Motion mask.}
The motion mask is derived from temporal changes in the generated samples. Given a sequence of spatiotemporal features or decoded frames \(\{Y_t\}_{t=1}^T\), we compute framewise motion maps using a temporal difference operator,
\begin{equation}
D_t = \phi(Y_t, Y_{t-1}),
\end{equation}
where \(\phi\) measures local temporal change. The resulting maps are normalized and projected onto the latent grid to obtain a latent saliency volume \(S_{\mathrm{motion}}\). Intuitively, this mask highlights regions with significant temporal variation, thereby emphasizing object motion, interaction changes, and potential dynamic inconsistencies.

\paragraph{Textaware mask.}
The textaware mask derives motion from cross-attention, making it explicitly prompt-grounded. For selected transformer layers and probe timesteps, we extract the cross-attention affinity between latent query tokens and prompt tokens, average over the relevant text tokens, and reshape the result into a spatiotemporal affinity map
\begin{equation}
A_t \in \mathbb{R}^{H \times W \times T}.
\end{equation}
We then convert this affinity into a motion-like signal by measuring its temporal variation,
\begin{equation}
S_{\text{text}}(t) = |A_t - A_{t-1}|,
\end{equation}
which highlights regions whose prompt-conditioned attention changes over time. In this way, the mask focuses on dynamic regions that are both visually active and semantically relevant to the text prompt.

\paragraph{Attention-based mask.}
The attention-based mask uses internal self-attention feature changes as a proxy for motion, similar in spirit to prior work using attention dynamics as an internal motion signal~\cite{pondaven2025video}. For selected self-attention layers, we extract latent query and key features at consecutive timesteps and measure their temporal similarity at each spatial location. Let \(q_t(h,w)\) and \(k_{t-1}(h,w)\) denote the corresponding features. We define the saliency as
\begin{equation}
S_{\text{attn}}(t,h,w) = \mathrm{ReLU}\!\left(1 - \cos\!\big(q_t(h,w), k_{t-1}(h,w)\big)\right).
\end{equation}
Locations whose internal features change more strongly over time therefore receive higher saliency. Unlike the other mask variants, this masking uses the model's own internal temporal feature evolution as a learned motion proxy.

\paragraph{Summary.}
The three masking strategies reflect different notions of dynamic saliency: the motion mask uses explicit visible motion, the textaware mask emphasizes prompt-grounded temporal changes, and the attention-based mask uses internal self-attention feature changes as a model-centric proxy for motion. As shown in Table~\ref{tab:masking_ablation}, all three provide useful dynamic signals, but induce different tradeoffs between semantic adherence and physical consistency.

\section{More Implementation Details}
\label{app:impl}

Unless otherwise stated, Proprio scoring is computed in latent space over a fixed set of scheduler timesteps sampled from the mid-noise regime, using multiple perturbation samples per timestep and inverse-variance weighting across timesteps. For refinement, we backpropagate through a reduced number of denoising steps while keeping the generator frozen, and optimize only the Gaussian-affine parameters of the initial noise. Our default motion score uses the motion-based spatiotemporal mask described in Sec.~\ref{sec:method}, which is fixed during scoring and refinement unless otherwise specified. For hybrid search-and-refinement, all methods use the same candidate pool before refinement.

Table~\ref{tab:opt_settings} summarizes the optimization hyperparameters used for Proprio self-refinement. Unless otherwise stated, we optimize the Gaussian-affine noise parameters for 10 steps using Adam with a learning rate of \(2\times 10^{-3}\). We apply gradient clipping for stability and use a small KL regularization term to keep the optimized noise close to the standard Gaussian prior. For score estimation during refinement, we average over 4 noise samples per timestep, which provides a more stable optimization signal while keeping the computational cost manageable. In all experiments, $\mu$ and $\sigma$ are calculated per channel. 
\begin{table}[h]
\centering
\small
\setlength{\tabcolsep}{6pt}
\caption{Optimization hyperparameters used for Proprio self-refinement.}
\label{tab:opt_settings}
\begin{tabular}{lc}
\toprule
\textbf{Hyperparameter} & \textbf{Value} \\
\midrule
Optimization steps ($K$) & 10 \\
Learning rate & $2 \times 10^{-3}$ \\
Gradient clipping & $3 \times 10^{-4}$ \\
KL regularization weight ($\beta$) & $5 \times 10^{-3}$ \\
Number of noise samples per timestep & 4 \\
\bottomrule
\end{tabular}
\end{table}

For all experiments, we use 4 mid-noise levels linearly spaced between 0.3-0.6 of the sampling steps. For non-distilled models, variance weighting is used by using 4 random iterations per noise level and calculating variance in prediction. For the hybrid loss, $\lambda = 0.2 $ is used for all models.

\section{Flow Matching Background} \label{app:flow_prelim}

Flow matching formulates generation as learning a time-dependent vector field that transports a simple base distribution to the data distribution in latent space. Let $z_1$ denote a clean video latent and let $z_0 \sim \mathcal{N}(0,I)$ denote a Gaussian noise sample. A continuous path between noise and data is defined by
\begin{equation}
z_t = (1-t) z_0 + t z_1, \qquad t \in [0,1],
\end{equation}
with corresponding target velocity
\begin{equation}
v^\star(z_0,z_1) = z_1 - z_0.
\end{equation}
A conditional generator $\hat{v}_\theta(z_t,t,c)$ is then trained to predict this transport field by minimizing
\begin{equation}
\mathcal{L}_{\mathrm{FM}}
=
\mathbb{E}_{z_1,z_0,t}
\left[
\left\|
\hat{v}_\theta(z_t,t,c) - (z_1-z_0)
\right\|_2^2
\right].
\end{equation}
In practice, modern video generators implement this perturbation through a scheduler rather than an explicit analytic path, so we write the noisy latent more generally as
\begin{equation}
z_t = \mathcal{P}(z_1,\epsilon,t), \qquad \epsilon \sim \mathcal{N}(0,I),
\end{equation}
and the corresponding target transport under the standard velocity parameterization as
\begin{equation}
v^\star(z_1,\epsilon) \approx \epsilon - z_1.
\end{equation}
At inference time, sampling starts from Gaussian noise and iteratively applies the learned vector field across timesteps to produce a latent video. In our method, we reuse this same flow-prediction objective at test time: by evaluating how well the frozen model predicts its target transport on perturbed generated samples, we obtain an intrinsic signal for ranking and refinement.

\section{Extended related work} \label{app:rel}
\paragraph{Physical plausibility in video generation.}
Recent video generative models~\cite{brooks2024sora,bar2024lumiere,kondratyuk2024videopoet,wan22,hunyuanvideo,yang2024cogvideox,hacohen2025ltxvideo,moviegen,physicsiq,kang2025far,physicsiq} have achieved impressive visual fidelity and temporal coherence, but multiple recent studies show that strong perceptual quality does not imply strong physical understanding or physically consistent dynamics. Diagnostic works such as Physics-IQ and related physical-law evaluations highlight persistent failures in collisions, object interactions, support relations, and generalization of physical behavior, suggesting that current video generators often reproduce surface realism without reliably capturing underlying physical structure~\cite{physicsiq,kang2025far,physicsiq}. Existing approaches to address this problem largely fall into three groups. A first line of work improves physical fidelity during training by introducing physics-aware data, supervision, or learning objectives, for example through physics-focused pretraining, preference optimization, or curated physics-aware corpora~\cite{wang2025wisa,cai2025phygdpo}. A second line of work incorporates explicit physical structure or simulation into the generation process itself. For example, PhysGen uses rigid-body simulation to ground image-to-video generation, while GPT4Motion leverages GPT-4 to script Blender-based physical motions before video synthesis~\cite{liu2024physgen,lv2024gpt4motion}. A third line of work addresses physical plausibility at inference time through external guidance. VLIPP uses a vision-language model as a coarse motion planner with physics-aware reasoning, and WMReward uses VJEPA-2 as an external latent world model to score candidates and guide sampling toward physically plausible videos~\cite{yang2025vlipp,wmreward}. Compared with these approaches, our method does not introduce external simulators, planners, or world-model rewards; instead, it reuses the frozen generator's own latent residual dynamics as a generator-native self-scoring signal for inference-time ranking and refinement.

\paragraph{Inference-time alignment for diffusion models.}
Inference-time alignment methods for diffusion models can be grouped into two broad families: \emph{search-based} methods and \emph{optimization-based} methods. Search-based methods improve alignment by spending additional compute on candidate exploration, trajectory branching, or resampling, followed by reward or value-based selection. This includes Best-of-$N$ style inference-time scaling, value-based decoding, and dynamic search or resampling procedures that explicitly trade off exploration and exploitation during denoising~\cite{svdd,dsearch,ma2025inferencescaling}. Optimization-based methods instead directly modify the sampling trajectory or the underlying noise variables to improve a target objective. In particular, DNO optimizes injected noise online during the denoising process with respect to a reward~\cite{dno}, while InitNO optimizes the initial latent using attention-derived prompt-alignment criteria and a distribution-preserving regularizer~\cite{initno}. More generally, training-free guidance methods steer diffusion trajectories at test time using external objectives without retraining the generator~\cite{ye2024tfg,askari2024cvs,dall2025chamfer}. Collectively, these works show that large gains can be achieved at inference time alone, without changing model weights. However, they typically depend on externally defined rewards, alignment heuristics, or verifier signals. In contrast, Proprio uses a generator-native objective: it scores and refines samples directly through the frozen video generator's own latent residual dynamics, avoiding external reward models.

\paragraph{Reward extraction from diffusion models.}
A broader line of work has explored whether diffusion models contain reward-like information that can be extracted or repurposed. In decision-making, \citet{nuti2023extracting} define a \emph{relative reward function} between a base and an expert diffusion model, and learn an auxiliary reward network by aligning its gradient to the difference in the two models' outputs. Their setting is primarily offline RL and control, but it provides an early indication that diffusion dynamics may encode preference-relevant structure beyond standard sample generation. 

\paragraph{Latent reward modeling for generation.}
More recent work has studied this question directly in image and video generation, arguing that rewards should be modeled in latent space rather than inherited from pixel-space VLM evaluators. Learning rewards directly on noisy latent states has been shown to better align with latent-space generation and reduce the overhead of external evaluation~\cite{dinalrm,videolrm}. Relatedly, LikePhys~\cite{likephys} evaluates intuitive physics understanding in video diffusion models by comparing denoising losses on curated valid--invalid simulator-generated pairs and summarizing the preference with a Plausibility Preference Error (PPE) metric. In contrast, Proprio does not train a separate reward model, does not learn from pairs of diffusion models, and does not require curated valid--invalid physics pairs. Instead, it directly reuses the frozen generator's own residual under controlled perturbations as a training-free inference-time self-scoring signal for ranking and refinement.

\section{Broader and societal impact.} \label{app:impact}
This work aims to improve the physical plausibility of generated videos, which may benefit applications such as robotics, simulation, and world modeling. More physically consistent video generation could make generative models more useful for studying agent behavior, planning, training embodied systems, and building controllable environments for decision-making and reasoning. At the same time, improving the realism of synthetic video also carries risks. More plausible generated videos may be harder to distinguish from real footage and could be misused for deception, misinformation, or other forms of misleading synthetic media. While we acknowledge these risks, our work is intended purely for academic research, and any examples produced in this work will be clearly labeled as synthetic. We hope this research contributes to safer and more controllable generative systems, while encouraging continued work on safeguards, transparency, and responsible deployment.

\section{Additional qualitative results}

\begin{figure}[H]
\centering
\includegraphics[width=\textwidth]{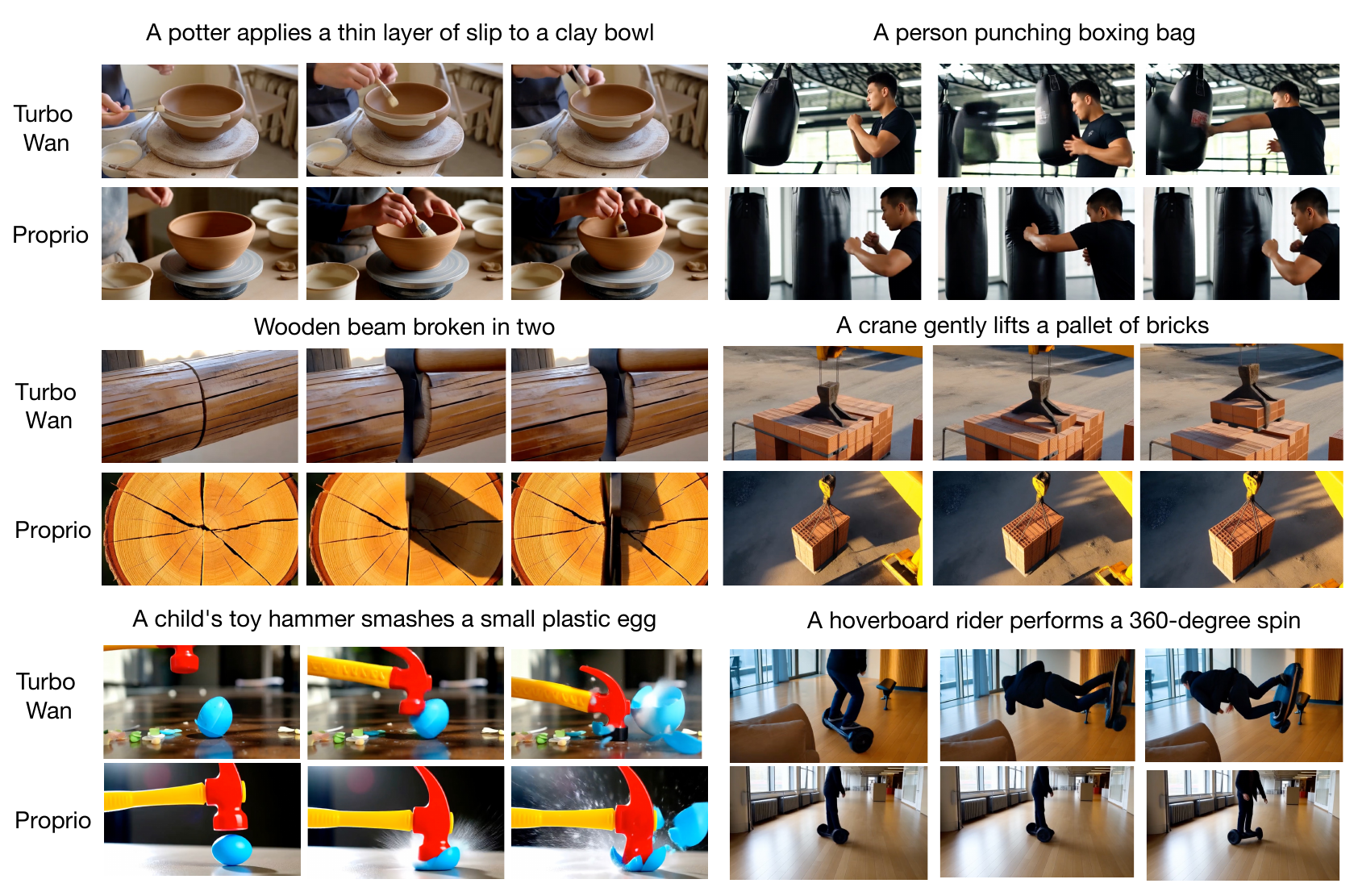}
 \caption{Overall qualitative results comparing our Proprio method against baseline TurboWan2.2. Visual examples from Proprio showing that our method produces more physically plausible generations for the T2V setting.}
\label{fig:qual1}
\end{figure}

\begin{figure}[H]
\centering
\includegraphics[width=\textwidth]{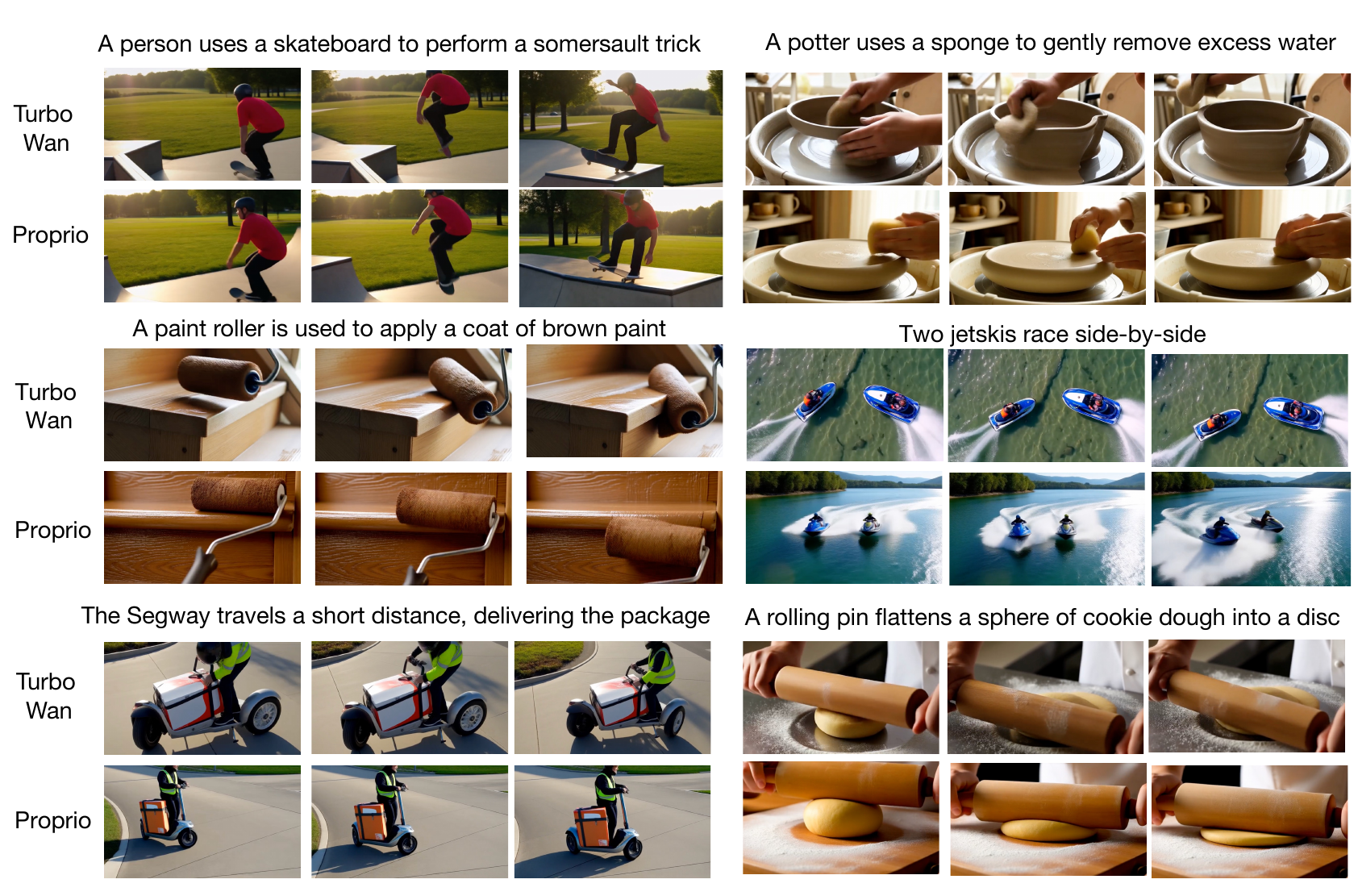}
 \caption{Overall qualitative results comparing our Proprio method against baseline TurboWan2.2. Visual examples from Proprio showing that our method produces more physically plausible generations for the T2V setting.}
\label{fig:qual2}
\end{figure}

\begin{figure}[H]
\centering
\includegraphics[width=\textwidth]{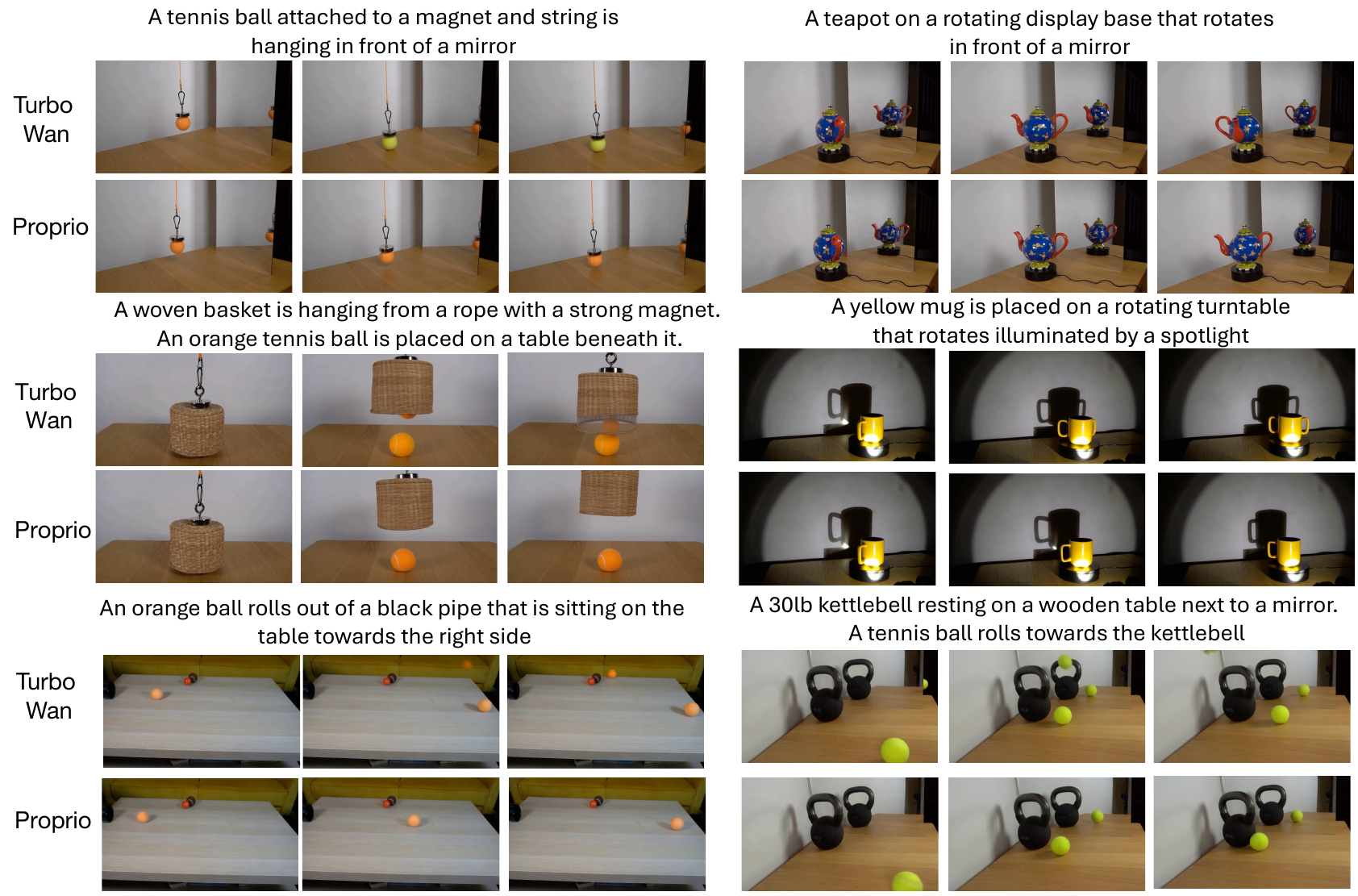}
 \caption{Overall qualitative results comparing our Proprio method against baseline TurboWan2.2. Visual examples from Proprio showing that our method produces more physically plausible generations for the I2V setting.}
\label{fig:qual3}
\end{figure}

\begin{figure}[H]
\centering
\includegraphics[width=\textwidth]{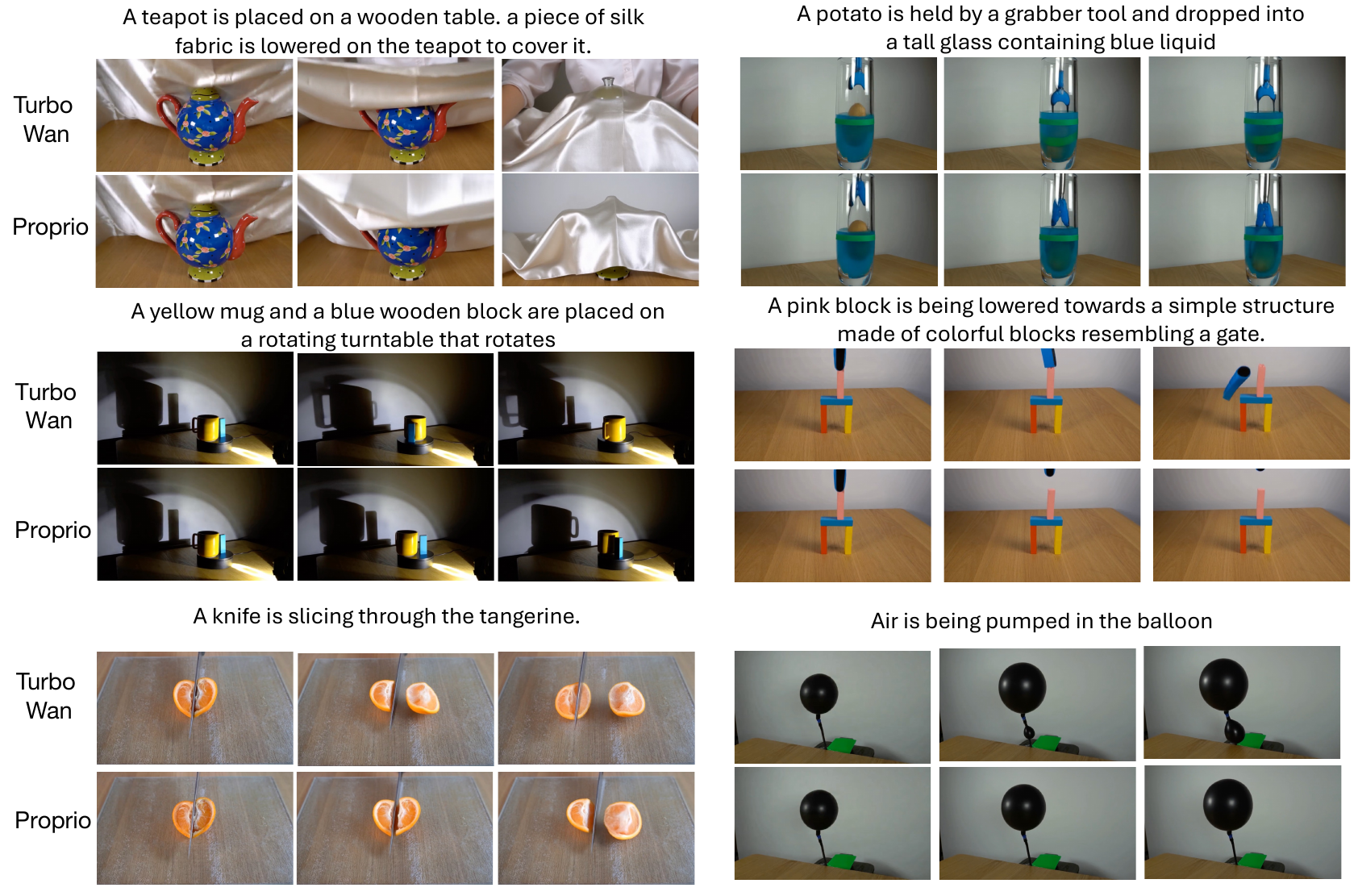}
 \caption{Overall qualitative results comparing our Proprio method against baseline TurboWan2.2. Visual examples from Proprio showing that our method produces more physically plausible generations for the I2V setting.}
\label{fig:qual4}
\end{figure}

Figures \ref{fig:qual1} and \ref{fig:qual2} as well as figures \ref{fig:qual3} and \ref{fig:qual4} show additional qualitative results for T2V and I2V settings respectively. We compare baseline performance on TurboWan2.2 with the optimized version highlighting the improvement in physics plausibility of the generations. We summarize the prompts used with the videos above the figures. We additionally provide full videos comparison in our website for better comparison on the dynamics as described in Appendix~\ref{app:demo}.
\newpage

\newpage

\end{document}